\documentclass{ecai} 
\usepackage{latexsym}
\usepackage{amssymb}
\usepackage{amsmath}
\usepackage{amsthm}
\usepackage{booktabs}
\usepackage{enumitem}
\usepackage{graphicx}
\usepackage{color}
\usepackage[lined,boxed,commentsnumbered,ruled]{algorithm2e}
\usepackage{subfigure}
\makeatletter
\renewcommand{\@thesubfigure}{\thesubfigure\space}
\makeatother
\usepackage{multirow}

\newcommand{\BibTeX}{B\kern-.05em{\sc i\kern-.025em b}\kern-.08em\TeX}

\begin{document}

%%%%%%%%%%%%%%%%%%%%%%%%%%%%%%%%%%%%%%%%%%%%%%%%%%%%%%%%%%%%%%%%%%%%%%%%

\begin{frontmatter}

%%% Use this command to specify your submission number.
%%% In doubleblind mode, it will be printed on the first page.

\paperid{9829} 

%%% Use this command to specify the title of your paper.

\title{Cascaded Large-Scale TSP Solving with Unified Neural Guidance: Bridging Local and Population-based Search}

%%% Use this combinations of commands to specify all authors of your 
%%% paper. Use \fnms{} and \snm{} to indicate everyone's first names 
%%% and surname. This will help the publisher with indexing the 
%%% proceedings. Please use a reasonable approximation in case your 
%%% name does not neatly split into "first names" and "surname".
%%% Specifying your ORCID digital identifier is optional. 
%%% Use the \thanks{} command to indicate one or more corresponding 
%%% authors and their email address(es). If so desired, you can specify
%%% author contributions using the \footnote{} command.

\author[A, B]{\fnms{Haoze}~\snm{Lv}}
\author[C]{\fnms{Wenjie}~\snm{Chen}} 
\author[A]{\fnms{Zhiyuan}~\snm{Wang}} 
\author[A, B]{\fnms{Shengcai}~\snm{Liu}\thanks{Corresponding Author. Email: liusc3@sustech.edu.cn}}
% \author[A]{\fnms{Ke}~\snm{Tang}\orcid{....-....-....-....}\thanks{Corresponding Author. Email: tangk3@sustech.edu.cn}} 

\address[A]{Guangdong Provincial Key Laboratory of Brain-Inspired Intelligent Computation, Department of Computer Science and Engineering, Southern University of Science and Technology, Shenzhen 518055, China}
\address[B]{Zhongguancun Academy, Beijing 100094, China}
\address[C]{Wenjie Chen, School of Information Management, Central China Normal University, Wuhan, China}

%%% Use this environment to include an abstract of your paper.

\begin{abstract}
The traveling salesman problem (TSP) is a fundamental NP-hard optimization problem. Over the past decades, traditional heuristic methods have achieved substantial success in solving TSP, yet their performance, particularly for large-scale instances, remains to be further improved. The advancement of deep learning technologies over the past decade has driven a growing number of attempts to solve TSP by leveraging neural guidance. However, these efforts predominantly focus on small-scale TSP instances, with limited improvements in solving performance for large-scale instances, revealing persistent scalability challenges.
This work presents UNiCS, a novel unified neural-guided cascaded solver for solving large-scale TSP instances.
UNiCS comprises a local search (LS) phase and a population-based search (PBS) phase, both guided by a learning component called unified neural guidance (UNG).
Specifically, UNG guides solution generation across both phases and determines appropriate phase transition timing to effectively combine the complementary strengths of LS and PBS. 
While trained only on simple distributions with relatively small-scale TSP instances, UNiCS generalizes effectively to challenging TSP benchmarks containing much larger instances (10,000-71,009 nodes) with diverse node distributions entirely unseen during training.
Experimental results on the large-scale TSP instances demonstrate that UNiCS consistently outperforms state-of-the-art methods, with its advantage remaining consistent across various runtime budgets.
\end{abstract}

\end{frontmatter}

%%%%%%%%%%%%%%%%%%%%%%%%%%%%%%%%%%%%%%%%%%%%%%%%%%%%%%%%%%%%%%%%%%%%%%%%

\section{Introduction}
\label{sec:intro}
The traveling salesman problem (TSP) is a fundamental NP-hard optimization problem with extensive applications~\cite{hubert1978applications,bland1989large,aoyama2004optimizing}.
Given a set of nodes and the distances between them, the goal of TSP is to find the shortest possible tour that visits each node exactly once and returns to the starting node.
Over the past decades, plenty of methods have been proposed to solve TSP.
Exact methods, including highly optimized ones like the Concorde solver~\cite{applegate2006concorde}, are guaranteed to find optimal solutions but suffer from worst-case exponential time complexity.
This limitation makes them impractical for solving large-scale TSP instances.
In contrast, heuristic search methods~\cite{LKH3:helsgaun2017extension}, although without guaranteeing optimality, can find high-quality solutions within reasonable computational time.

In general, most (if not all) traditional heuristic methods are designed based on expert knowledge, making them human-interpretable.
However, with the advancement of deep learning technologies over the past decade, there has been growing interest in training powerful deep neural networks to solve TSP instances generated from specific distributions~\cite{PointerNetwork:conf/nips/VinyalsFJ15,AM:conf/iclr/KoolHW19}.
Unfortunately, despite continuous improvements in performance, these deep learning models still fall significantly behind traditional heuristic methods in terms of solution quality~\cite{liu2023good}.
This performance gap becomes particularly evident when dealing with large-scale TSP instances containing thousands of nodes and real-world TSP instances that may not conform to the distributions used during training.

% enhance local search & 其本质上的缺陷
Rather than training purely neural networks to solve TSP, combining strong traditional heuristic methods with learning techniques is believed to be a more practical approach for advancing the state-of-the-art TSP solving capabilities, as highlighted by~\cite{BengioLP21}.
Through learning from experience, the learning components, which are integrated into the heuristic methods, aim to discover better policies to replace hand-crafted rules for making critical algorithmic decisions during the search process.
Throughout this paper, these methods are refereed to as hybrid methods.
Representative hybrid methods for solving TSP include NLKH~\cite{Neurolkh:conf/nips/XinSCZ21} and VSR-LKH~\cite{VSR-LKH:journals/kbs/ZhengHZJL23} that enhance the powerful Lin-Kernighan-Helsgaun (LKH) method~\cite{LKH3:helsgaun2017extension} by learning policies for candidate edge generation and selection during LKH's $\lambda$-opt search process, respectively.
Both NLKH and VSR-LKH have shown considerable improvements over the original LKH method across a wide range of TSP instances.
However, they are still inherently limited by the underlying local search (LS) mechanism of LKH.
In general, LS methods often exhibit limited global search capabilities and tend to get trapped in local optima.
This becomes particularly problematic for large-scale TSP instances, where the presence of numerous local optima can significantly impact the solution quality obtained by LS methods such as LKH and its hybrid variants.

Besides LS, population-based search (PBS) like genetic algorithm (GA) represents another powerful heuristic framework for solving TSP.
By maintaining and updating a population of solutions, PBS typically exhibits stronger capabilities in exploring the solution space compared to LS methods.
Taking advantage of this, powerful PBS methods such as GA with edge assembly crossover (EAX)~\cite{EAX:journals/informs/NagataK13}  and multi-agent optimization system (MAOS)~\cite{XieL09}
can achieve better solution quality than LS methods like LKH, when solving TSP instances with more than 10,000 nodes.
However, due to the computational overhead from evolving many solutions simultaneously, PBS methods suffer from slower convergence -- they cannot yield high-quality solutions as rapidly as LS methods.

This work aims to boost the state-of-the-art in solving large-scale TSP instances involving more than 10,000 nodes.
The core idea is to develop a learning-based guidance that directs a hybrid solver combining the complementary strengths of LS and PBS.
Specifically, a novel \textbf{u}nified \textbf{n}eural-gu\textbf{i}ded \textbf{c}ascaded \textbf{s}olver, dubbed UNiCS, is proposed that implements a cascaded search process consisting of two phases: a LS phase following the LKH framework, and a PBS phase following the EAX framework.
The cascaded mechanism leverages the rapid convergence of LS phase to identify high-quality solutions, which are then further improved through the exploration capabilities of PBS phase.
Both phases are guided by a unified neural guidance (UNG) module in two critical aspects.
First, it guides the generation of high-quality solutions in both phases by scoring edges to evaluate their likelihood of appearing in optimal solutions.
Second, it determines the appropriate timing for transitioning from LS phase to PBS phase.
This transition timing varies with problem size and can significantly improve UNiCS's overall performance.

Unlike many existing deep learning models that focus on solving TSP instances  from specific distributions, UNiCS is evaluated under more challenging settings.
Trained only on simple distributions with relatively small-scale instances, UNiCS is tested on conventional TSP benchmarks TSPlib~\cite{reinelt1991tsplib}, National~\cite{National_TSPdata}, and VLSI~\cite{VLSI_TSPdata} that differ substantially from the training data.
These benchmarks contain much larger instances (with 10,000 to 71,009 nodes) and feature diverse node distributions entirely unseen during training.
Experimental results on these instances show that UNiCS achieves superior solution quality compared to state-of-the-art heuristic methods and hybrid methods with learning components.
Moreover, the performance advantage of UNiCS remains consistent under various runtime budgets, demonstrating its strong practicality to different real-world time constraints.
\section{A Brief Review on Methods for Solving TSP}
\label{sec:related_work}

\subsection{Traditional Methods}
\label{sec:traditional_method}
Traditional methods for solving TSP can be broadly classified into exact and heuristic ones.
Exact methods like Concorde~\cite{Concorde:journals/orl/Trick08} utilize branch-and-cut techniques to find optimal solutions, but are impractical for large-scale instances due to exponential time complexity.
Heuristic methods, on the other hand, aim to find near-optimal solutions efficiently.
These can be further divided into LS-based and PBS-based ones~\cite{gutin2006traveling,XieL09}, with LKH~\cite{LKH3:helsgaun2017extension} and EAX~\cite{EAX:journals/informs/NagataK13} being representative methods of each category respectively.
LKH adopts iterative local search with $\lambda$-opt moves and $\alpha$-nearness measure to reduce search space, offering quick high-quality solutions but tends to get trapped in local optima.
EAX maintains a population of solutions and evolves them through sophisticated edge assembly crossover, showing strong performance particularly on large-scale instances.
Since EAX forms the foundation for the PBS phase in UNiCS, its key components are introduced below (see Section~C in the supplementary material\cite{liu2025cascaded} for a detailed description of EAX).
While LKH similarly forms the foundation for the LS phase in UNiCS, its implementation under UNG aligns with NLKH and can be found in Section~D in the supplementary material\cite{liu2025cascaded}.

The search process of EAX is divided into two stages: Stage I for the early and middle phases, and Stage II for the final phase.
The core of EAX lies in its edge assembly crossover that generates offspring solutions from two selected parent solutions. 
Let $V$ be the set of all nodes of a TSP instance.
Given parent solutions $p_A$ and $p_B$, let $E_A$ and $E_B$ be the sets of edges included in $p_A$ and $p_B$, respectively, and let $N_{ch}$ be the number of offspring to be generated.
The crossover proceeds as follows:
\begin{itemize}
\item \textbf{Step 1:} A multigraph $G_{AB}=(V,E_A \cup E_B)$ is first constructed and then partitioned into $AB$-cycles through random walks on the multigraph, where each $AB$-cycle contains edges from $E_A$ and $E_B$ in alternating sequence.
\item \textbf{Step 2:} An $AB$-cycle is selected based on a predefined strategy, and the set of its edges forms the $E$-set.
\item \textbf{Step 3:} An intermediate solution is created by modifying $p_A$. This modification involves removing edges in $E_A \cap E\text{-set}$ from $E_A$ and adding edges in $E_B \cap E\text{-set}$ to $E_A$.
The resulting intermediate solution may contain one or more sub-tours and thus be invalid. These sub-tours are then combined through a specific merging process to create a valid solution.
\item \textbf{Step 4:} Steps 2-3 are repeated until $N_{ch}$ offspring are generated.
\end{itemize}

The $AB$-cycle selected in step 2 directly determines the $E$-set and the resulting offspring, making it the crucial part of the EAX method.
Indeed, various selection strategies were designed and evaluated in~\cite{EAX:journals/informs/NagataK13}, and the final adopted approach in EAX was random selection in Stage I and Tabu search in Stage II.
In the PBS phase of our proposed UNiCS, a learning component (UNG) replaces these hand-crafted rules to make better $AB$-cycle selections.

\begin{figure*}[tbp]
\centering
\includegraphics[width=\textwidth]{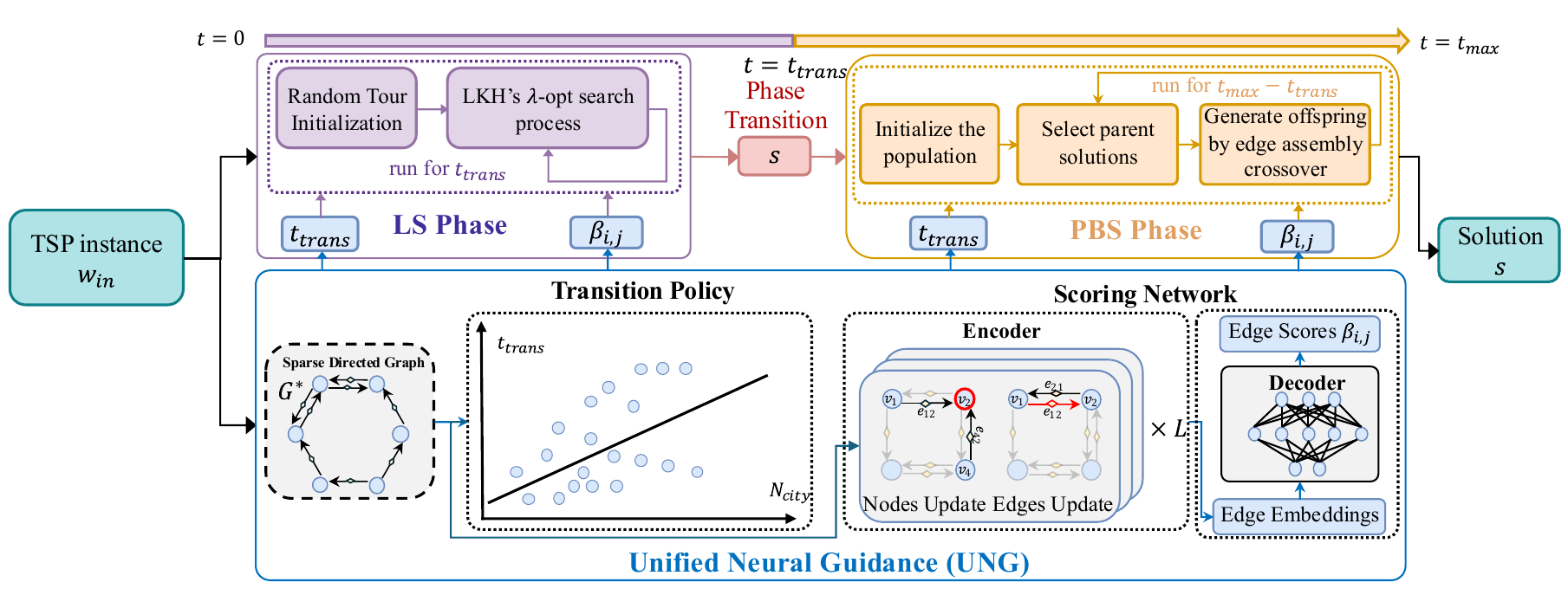} % Reduce the figure size so that it is slightly narrower than the column. Don't use precise values for figure width.This setup will avoid overfull boxes.
\vspace{-5mm}
\caption{An overview of UNiCS.}
\label{fig:UNiCS}
\end{figure*}

\subsection{Deep Learning Models}
Deep learning models for solving TSP can be categorized into constructive and improvement ones.
The former build solutions sequentially, starting with Pointer Network~\cite{PointerNetwork:conf/nips/VinyalsFJ15} and advancing through reinforcement learning~\cite{BelloRL:conf/iclr/BelloPL0B17} and Transformer-based architectures~\cite{AM:conf/iclr/KoolHW19}. 
Recent works have explored employing solution symmetry~\cite{POMO:conf/nips/KwonCKYGM20}, policy ensemble~\cite{gao2023towards}, and divide-and-conquer strategies~\cite{H-TSP:conf/aaai/PanJDF0S023,GLOP:conf/aaai/YeWLCLL24} to tackle larger TSP instances.
Improvement models focus on improving existing solutions, starting with controlling 2-opt operator and advancing to novel node embedding~\cite{ma2021learning} and node pair selection~\cite{wu2021learning}.
Despite the ongoing progress, deep learning models still significantly lag behind heuristic methods in solution quality, especially for large-scale TSP instances.
For more details on this direction, one may refer to the recent survey~\cite{cappart2023combinatorial}.

\subsection{Hybrid Methods}
Hybrid solvers combining heuristic methods with learning techniques have shown promise for large-scale TSP instances.
VSR-LKH~\cite{VSR-LKH:journals/kbs/ZhengHZJL23} used reinforcement learning for edge selection during the $\lambda$-opt process of LKH, while NLKH~\cite{Neurolkh:conf/nips/XinSCZ21} employed a graph neural network to quickly generate high-quality candidate edge sets for LKH. 
Built upon VSR-LKH, RHGA~\cite{RHGA:journals/cor/ZhengZCH23} combined it with EAX, using the learned Q-values to guide tour generation in EAX. In contrast to UNiCS, RHGA adopts a memetic algorithm framework, where VSR-LKH is applied to further refine the offspring population generated in each iteration of EAX.
Beyond LKH-based methods, enhancing other types of heuristic methods such as the ant colony optimizer~\cite{DeepACO:conf/nips/YeWCLL23} has also been explored.
Another research direction focuses on incorporating different TSP heuristic methods into algorithm portfolios, where methods either run independently in parallel~\cite{LiuT019,TangLYY21,LiuTY22} or are selected based on machine learning models~\cite{zhao2021towards} when solving problem instances.
Our proposed solver, UNiCS, differs from existing hybrid solvers in that UNG directs and coordinates both LS and PBS phases throughout the solving process, which is the first attempt of such integration in the literature.

\section{A Unified Neural-Guided Cascaded Solver}
As shown in Figure~\ref{fig:UNiCS}, UNiCS consists of two cascaded phases: a LS phase following the LKH framework and a PBS phase following the EAX framework.
Both phases are guided by the UNG module.
High-quality solutions discovered during the LS phase are incorporated into the initial population of the PBS phase to leverage the strengths of both search paradigms.
The role of UNG in guiding solution generation and determining the phase transition timing is first described below.
Then, the complete UNiCS solver is presented.

\subsection{Solution Generation Guided by UNG}
\label{sec:solution_generation}
UNG contains a neural network that scores edges to estimate their likelihood of appearing in optimal solutions.
This scoring network guides solution generation in both LS and PBS phases.
For the LS phase, the scoring network guides candidate edge generation during LKH's $\lambda$-opt process.
This implementation aligns with NLKH and is thus omitted here due to space limitation (details provided in Section~D in the supplementary material\cite{liu2025cascaded}).
For the PBS phase, the scoring network guides $AB$-cycle selection during EAX's crossover process, which is detailed below.

\vspace{+0.5mm}
\noindent\textbf{Scoring Network.} Inspired by NLKH~\cite{Neurolkh:conf/nips/XinSCZ21}, the scoring network is implemented as a sparse graph network (SGN) to handle large-scale TSP instances efficiently.
The input TSP instance is first converted to a sparse directed graph $G^* = (V, E^*)$, where $V$ represents the set of nodes and $E^*$ contains only the $\gamma$-shortest edges originating from each node.
% Note that an edge $(i, j)$ belonging to $E^*$ does not necessarily mean that the reverse edge $(j, i)$ also belongs to $E^*$.
The SGN consists of an encoder that embeds edges and nodes into feature vectors, as well as a decoder used for edge scoring.
The encoder first linearly projects the node input $x_v \in V$ and the edge input $x_e \in E^*$ into feature vectors $v^0_i \in R^D$ and $e^0_{i,j} \in R^D$, where $D$ is the feature dimension; then the node and edge features are recursively embedded with $L$-layer SGN:
\begin{equation}
\label{eq1}
    a_{i,j}^l = \exp \left( W_a^l e_{i,j}^{l-1} \right) \oslash \sum_{(i,k) \in E^*} \exp \left( W_a^l e_{i,k}^{l-1} \right),
\end{equation}
\begin{equation}
\label{eq2}
   v_i^l = \mathcal{F} \left( W_s^l v_i^{l-1} + \sum_{(i,j) \in E^*} a_{i,j}^l \odot W_n^l v_j^{l-1} \right) + v_i^{l-1},
\end{equation}
\begin{equation}
\label{eq3}
       r_{i,j}^l = \begin{cases}
    W_r^l e_{j,i}^{l-1}, & \text{if } (j,i) \in E^* \\
    W_r^l p, & \text{otherwise}
    \end{cases},
\end{equation}
\begin{equation}
\label{eq4}
     e_{i,j}^l = \mathcal{F} \left( W_f^l v_i^{l-1} + W_t^l v_j^{l-1} + W_o^l e_{i,j}^{l-1} + r_{i,j}^l \right) + e_{i,j}^{l-1}.
\end{equation}
$W^l_a, W^l_n, W^l_s, W^l_r, W^l_f, W^l_t, W^l_o \in R^{D\times D}$ are trainable parameters, with $l = 1, 2, \dots, L$ being the layer index.
$\odot$ and $\oslash$ represent element-wise multiplication and element-wise division, respectively.
$\mathcal{F}$ represents ReLU activation followed by Batch Normalization.

Based on $e^L_{i,j}$ output by the encoder, the decoder generates the final embedding vectors $e^F_{i,j}$ using two layers of linear projection and ReLU activation.
Then, the edge score $\beta_{i,j}$ is calculated as follows:
\begin{equation}
\label{eq5}
    \beta_{i,j} = \frac{\exp \left( W_\beta e_{i,j}^F \right)}{\sum_{(i,k) \in E^*} \exp \left( W_\beta e_{i,k}^F \right)},
\end{equation}
where $W_\beta \in R^{D} $ are trainable parameters.
SGN is trained by supervised learning with the followig loss:
\begin{equation}
\label{eq6}
    \begin{aligned}
        \mathcal{L}_\beta = -\frac{1}{\gamma \left\vert V\right\vert} \sum\limits_{(i,j)\in E^*} ( & \mathbb{I}\left\{(i,j) \in E_o^*\right\}\log\left(\beta_{i,j}\right) \\
       & + \mathbb{I} \left\{(i,j) \notin E_o^*\right\} \log (1-\beta_{i,j})),
    \end{aligned}
\end{equation}
where $\mathbb{I}$ is the indicator function and $E^*_o$ is the optimal solution to the training TSP instance.
Once trained sufficiently, $\beta_{i,j}$ would measure the likelihood of edge $(i, j)$ appearing in the optimal solution.

\vspace{+0.5mm}
\noindent\textbf{Evaluation of $AB$-Cycles.} The original $AB$-cycle selection in EAX relies on hand-crafted rules.
In contrast, in the PBS phase of UNiCS, the learned scoring network is employed to evaluate $AB$-cycles to guide the selection process.
As explained in Section~\ref{sec:traditional_method}, EAX creates intermediate solutions by modifying the parent solution $p_A$: removing edges from $E_A$ within the $E$-set (formed by the selected $AB$-cycle) and adding edges from $E_B$.
This means $AB$-cycle edges from $E_A$ are removed in the solution, while those from $E_B$ are included.
Therefore, the quality of an $AB$-cycle is evaluated using Eq.~(\ref{eq7}):
\begin{equation}
\label{eq7}
    \text{score}_{AB} = \sum_{(i,j) \in E_B \cap E_{AB}} \beta_{i,j} - \sum_{(i,j) \in E_A \cap E_{AB}} \beta_{i,j},
\end{equation}
where $\text{score}_{AB}$ is the score of a $AB$-cycle, and $E_{AB}$ is its edge set.
Specifically, Eq.~(\ref{eq7}) sum the scores of edges in $E_B \cap E_{AB}$, as they will be added to the offspring, and subtract the scores of edges in $E_A \cap E_{AB}$, as they will be removed.
This approach ensures that the final score reflects the potential benefit of including edges from $E_B$ while considering the cost of excluding edges from $E_A$.

\vspace{+0.5mm}
\noindent\textbf{Selection of $AB$-Cycles.}
Based on the evaluation approach described above, a selection strategy is developed to leverage these scores effectively.
For each generated $AB$-cycle, its $\text{score}_{AB}$ is first evaluated using Eq.~(\ref{eq7}).
In step 2 of the EAX crossover process (see Section~\ref{sec:traditional_method}), $AB$-cycles are selected sequentially in descending order of their scores to generate offspring.
On the other hand, relying solely on scores could lead to a loss of diversity among offspring.
To mitigate this issue, each selection has a probability $\eta$ of randomly choosing an unused $AB$-cycle instead of following the score-based order, where $\eta \in (0,1)$ is a hyperparameter.

\begin{algorithm}[tbp]
\LinesNumbered
\KwIn{TSP Instance $w_{in}$, runtime budget $t_{\max}$}
\KwOut{TSP solution $s$}
$G^*\leftarrow$ Convert $w_{in}$ to a sparse directed graph;\\
$\beta_{i,j}, t_{trans} \leftarrow \mathrm{UNG}(G^*)$;\\
$s \leftarrow \text{LS}(w_{in}, \beta_{i,j}, t_{trans})$;\\
$s \leftarrow \text{PBS}(w_{in}, s, \beta_{i,j}, t_{max} - t_{trans})$;\\
\Return{$s$}
\caption{UNiCS}
\label{alg:UNiCS}
\end{algorithm}

\begin{figure*}[h]
\centering
\includegraphics[width=0.9\textwidth]{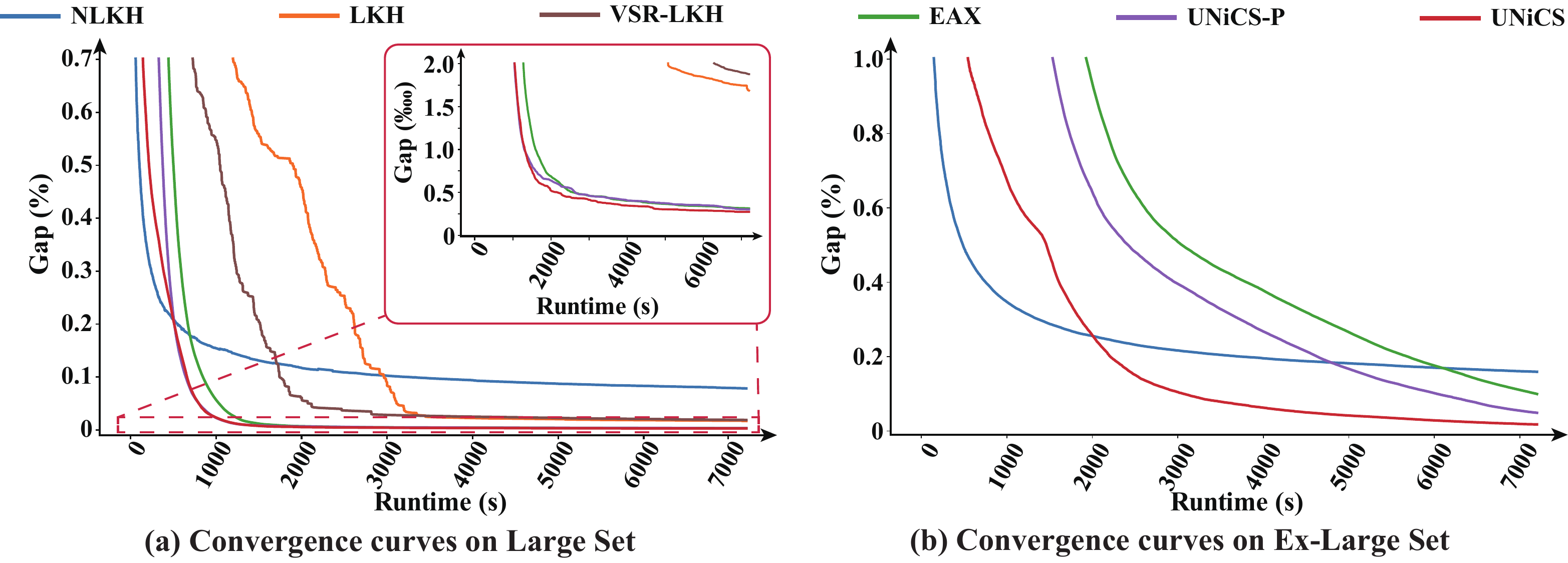}
\caption{Convergence curves averaged over 10 runs.}
\vspace{-3mm}
\label{fig2}
\end{figure*}

\subsection{Phase Transition Guided by UNG}
\label{sec:phase_transition}
A key observation is that the transition time $t_{trans}$ between the LS and PBS phases (see Figure~\ref{fig:UNiCS}) is crucial to UNiCS's performance.
The LS phase exhibits efficient early-stage search capabilities but may get trapped in local optima, while the PBS phase shows stronger exploration capabilities but converges more slowly.
The transition time $t_{trans}$ represents both the runtime allocated to the LS phase and the runtime deducted from the PBS phase's total budget.
If $t_{trans}$ is set too short, the rapid convergence capability of the LS phase cannot be fully exploited.
Conversely, if $t_{trans}$ is set too long, the search process may enter the LS phase's relatively inefficient stage where it becomes trapped in local areas, potentially leading to diminishing returns.
Therefore, determining an appropriate $t_{trans}$ is critical.
Unfortunately, the optimal value of $t_{trans}$ is unknown beforehand and varies across different TSP instances.

On the other hand, it has been widely observed that the complementary strengths of LS and PBS manifest distinctly across TSP instances of different sizes~\cite{EAX:journals/informs/NagataK13,zhao2021towards}.
This observation suggests that problem size could serve as a crucial factor in determining the appropriate $t_{trans}$. 
Based on this insight, a transition policy mapping from problem size to $t_{trans}$ is established within the UNG module.
The policy takes the sparse graph $G^*$ of a TSP instance as input and outputs the corresponding $t_{trans}$.
To train this policy, random uniform TSP instances of various sizes are used as training instances, with UNiCS being run on them under different $t_{trans}$ settings to gather training data (detailed in Section~\ref{sec:exp_setup}).
For each training instance, different $t_{trans}$ settings are compared by evaluating the area under the optimality gap curve (detailed in Appendex~B).
This metric provides a comprehensive assessment of UNiCS's performance across the entire runtime rather than at specific runtime points, and its aggregated nature also helps mitigate the impact of randomness in performance assessment.
Then, the optimal $t_{trans}$ with the smallest area is identified. 
After obtaining training data, a linear model is fitted to these optimal values to derive the transition policy.
This simple model is chosen to avoid overfitting while maintaining interpretability and computational efficiency.
UNG then uses the trained policy to automatically determine $t_{trans}$ for testing instances.

\subsection{The Complete Solver: UNiCS}
The complete UNiCS solver integrates UNG with both LS and PBS phases, as outlined in Algorithm~\ref{alg:UNiCS}. 
Given a TSP instance $w_{in}$ and runtime budget $t_{max}$, UNiCS starts by converting $w_{in}$ to a sparse directed graph $G^*$ (line 1) that serves as input to UNG.
UNG then process this graph to generates edge scores $\beta_{i,j}$ and transition time $t_{trans}$ (line 2).
During the LS phase, these edge scores guide the candidate edge generation in LKH's $\lambda$-opt process until $t_{trans}$ is reached (line 3).
After that, the solver transitions to the PBS phase.
The best solution obtained from the LS phase is incorporated into the initial population by replacing a randomly selected solution.
The PBS phase then continues the search for the remaining runtime $t_{max} - t_{trans}$, with edge scores guiding $AB$-cycle selection as described earlier (line 4).
Finally, UNiCS returns the best found solution (line 5).

\section{Computational Studies}
\label{sec:exp}

\begin{table*}[h]
\centering
\caption{Relative gap (\%) to the BKS on the Large set (10 runs per instance). ``-'' in Best / Avg columns indicates no solutions produced in 10 runs / solutions produced only in some runs. Bold indicates best performance. ``Total'' row summarizes average gaps across Large set.} 
\label{table1}
\resizebox{\textwidth}{!}{
\begin{tabular}{cccccccccccccc}
\toprule
\multirow{2}{*}{Runtime} & \multirow{2}{*}{Benchmark} & \multicolumn{2}{c}{NLKH} & \multicolumn{2}{c}{LKH} & \multicolumn{2}{c}{VSR-LKH} & \multicolumn{2}{c}{EAX} & \multicolumn{2}{c}{UNiCS-P} & \multicolumn{2}{c}{UNiCS} \\ \cmidrule(lr){3-4} \cmidrule(lr){5-6} \cmidrule(lr){7-8} \cmidrule(lr){9-10} \cmidrule(lr){11-12} \cmidrule(lr){13-14}
& & Best & Avg & Best & Avg & Best & Avg & Best & Avg & Best & Avg & Best & Avg \\ \midrule
\multirow{4}{*}{1800s} & National & 0.0922 & 0.1384 & - & - & 0.0072 & - & 0.0067 & 0.0124 & 0.0053 & 0.0063 & \textbf{0.0049} & \textbf{0.0063} \\
& VISI & 0.0958 & 0.1360 & - & - & 0.0260 & - & 0.0053 & 0.0078 & 0.0052 & 0.0077 & \textbf{0.0045} & \textbf{0.0066} \\
& TSPLib & 0.0174 & 0.0567 & 0.0017 & 0.0082 & 0.0058 & 0.0127 & \textbf{0.0007} & 0.0014 & 0.0008 & 0.0022 & 0.0010 & \textbf{0.0013} \\
& Total & 0.0852 & 0.1266 & - & - & 0.0193 & - & 0.0050 & 0.0080 & 0.0045 & 0.0066 & \textbf{0.0041} & \textbf{0.0059} \\ \midrule
\multirow{4}{*}{3600s} & National & 0.0666 & 0.0965 & 0.0019 & 0.0117 & 0.0048 & 0.0099 & 0.0019 & 0.0027 & \textbf{0.0019} & 0.0029 & 0.0023 & \textbf{0.0029} \\
& VISI & 0.0774 & 0.1096 & 0.0133 & 0.0314 & 0.0175 & 0.0353 & 0.0037 & 0.0053 & 0.0040 & 0.0055 & \textbf{0.0031} & \textbf{0.0044} \\
& TSPLib & 0.0145 & 0.0446 & 0.0015 & 0.0055 & 0.0018 & 0.0059 & 0.0007 & 0.0011 & \textbf{0.0005} & \textbf{0.0008} &\textbf{0.0005} & 0.0010 \\
& Total & 0.0672 & 0.0986 & 0.0093 & 0.0239 & 0.0128 & 0.0261 & 0.0029 & 0.0042 & 0.0031 & 0.0043 & \textbf{0.0026} & \textbf{0.0037} \\ \midrule
\multirow{4}{*}{7200s} & National & 0.0484 & 0.0693 & 0.0011 & 0.0044 & 0.0028 & 0.0066 & 0.0015 & 0.0020 & 0.0016 & 0.0021 & \textbf{0.0016} & \textbf{0.0019} \\
& VISI & 0.0611 & 0.0914 & 0.0084 & 0.0236 & 0.0116 & 0.0258 & 0.0027 & 0.0040 & 0.0026 & 0.0038 & \textbf{0.0022} & \textbf{0.0034} \\
& TSPLib & 0.0118 & 0.0255 & 0.0001 & 0.0036 & 0.0007 & 0.0033 & 0.0004 & 0.0005 & \textbf{0.0002} & \textbf{0.0004} & 0.0003 & 0.0006 \\
& Total & 0.0521 & 0.0783 & 0.0057 & 0.0169 & 0.0083 & 0.0188 & 0.0022 & 0.0031 & 0.0021 & 0.0030 & \textbf{0.0018} & \textbf{0.0027} \\
\bottomrule
\end{tabular}
}
\end{table*}

\begin{figure*}[tb]
\centering
\includegraphics[width=1.0\textwidth]{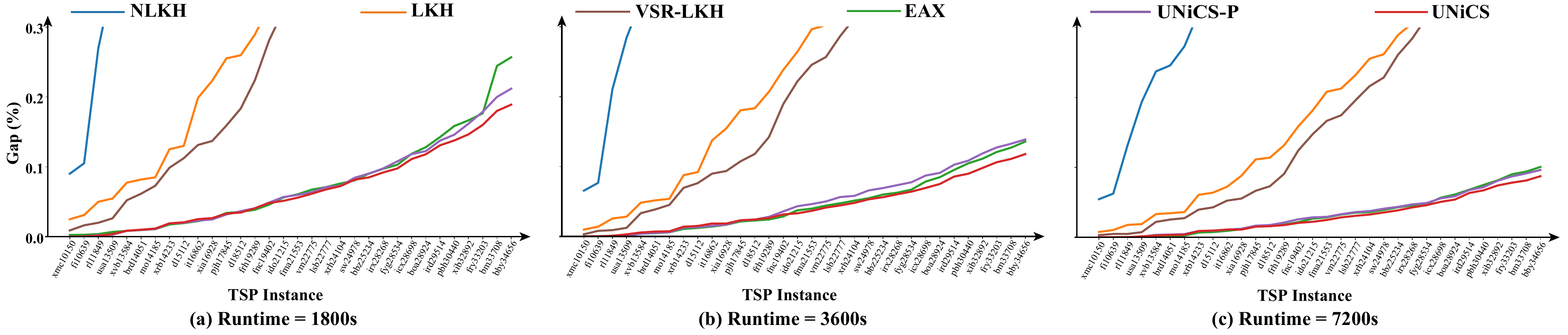}
\vspace{-3mm}
\caption{Cumulative gaps on Large set with instances sorted by ascending problem size.}
\label{fig3}
\end{figure*}

\begin{figure}[h]
\centering
\includegraphics[width=\columnwidth]{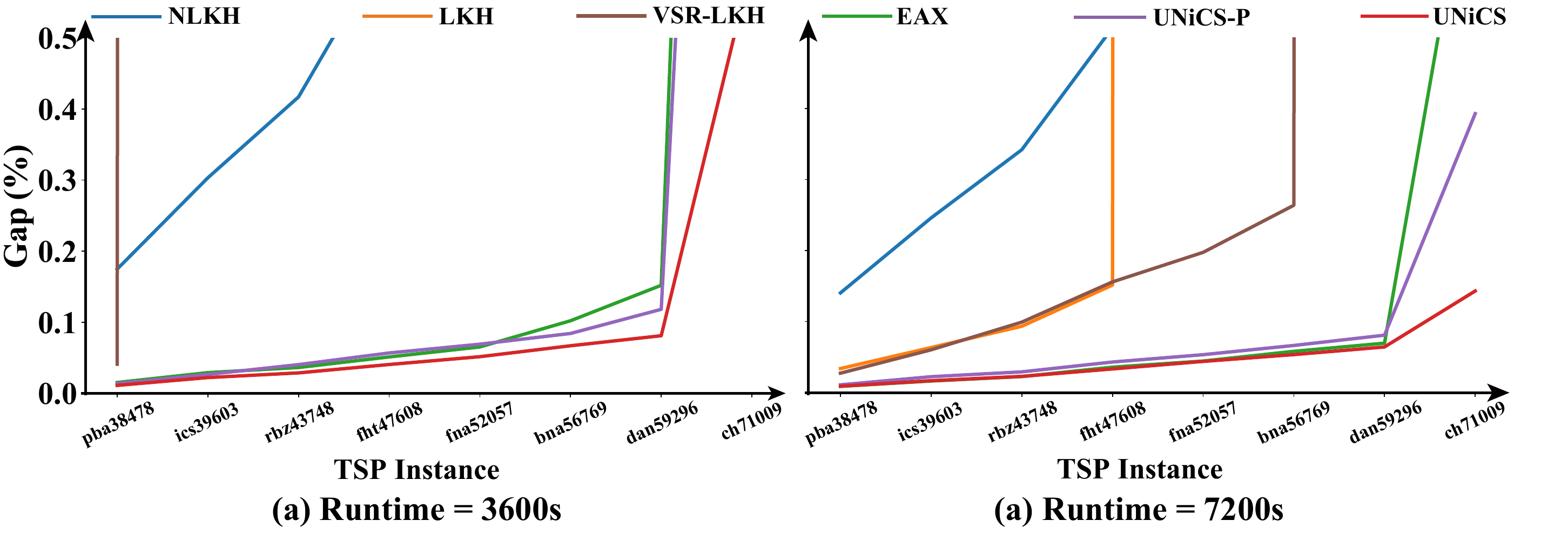}
\vspace{-3mm}
\caption{Cumulative gaps on Ex-Large set with instances sorted by ascending problem size.}
\label{fig4}
\end{figure}

Unlike existing deep learning models that are typically evaluated on TSP instances generated from specific distributions, UNiCS is tested under more challenging and realistic settings to thoroughly assess its capabilities.
Specifically, UNiCS is trained only on random uniform instances (which can be easily obtained in practice), and is tested on diverse TSP benchmarks that feature much larger problem sizes and significantly different node distributions from the training data.
Moreover, the testing is conducted across different runtime budgets to examine UNiCS's performance under varying time constraints.
Through these experiments, two key research questions (RQs) are investigated.
\textbf{RQ1:} Can UNiCS effectively generalize to these challenging benchmarks? Specifically, how does it perform compared to state-of-the-art heuristic and hybrid methods under different runtime budgets?
\textbf{RQ2:} Are the $AB$-cycle selection and phase transition guided by UNG contribute meaningfully to UNiCS's performance?

\subsection{Experimental Setup}
\label{sec:exp_setup}
\noindent\textbf{Training UNG.}
The scoring network SGN within UNG was trained on two-dimensional Euclidean TSP instances with 500 to 1000 nodes, where node coordinates were uniformly sampled from a unit square.
Since the number of optimal edges for each instance is proportional to the number of nodes, $\frac{500000}{V}$ instances  were generated for each problem size $V$ in the training set, resulting in approximately 346,000 training instances in total.
The optimal edges $E^*_o$ for each instance were obtained using Concorde.
Each instance was converted to a sparse directed graph with $\gamma = 20$ edges per node (see Section~\ref{sec:solution_generation}).
The SGN consists of $L=25$ sparse graph convolution layers with a hidden dimension $D = 128$ and 4 attention heads.
The model was trained for 100 epochs using Adam optimizer with a learning rate of $10^{-4}$.

For the transition policy within UNG, 56 TSP instances ranging from 3,000 to 30,000 nodes were generated using the same uniform distribution described above.
For each instance, the transition time was varied from 50 to 650 seconds in 50-second increments.
The optimal $t_{trans}$ was then determined by evaluating the area under the optimality gap curve (detailed in Section~B in the supplementary material\cite{liu2025cascaded}).

\vspace{+0.5mm}\noindent\textbf{TSP Benchmarks.}
All two-dimensional Euclidean TSP instances with more than 10,000 nodes were collected from three TSP benchmark sets: TSPLib~\cite{reinelt1991tsplib}, National~\cite{National_TSPdata}, and VLSI~\cite{VLSI_TSPdata}, totaling 38 instances.
These benchmarks represent diverse real-world scenarios.
VLSI instances are derived from integrated circuit applications, National instances are based on city distances in different countries, while TSPLib includes various sources such as logistics, circuit board drilling, and transistor routing.
The collected instances were further divided into two sets: Large set (32 instances, 10,000-40,000 nodes) and Ex-Large set (6 instances, 40,000+ nodes).
The Large set was further categorized by source: VLSI (21 instances), National (7 instances), and TSPLib (4 instances).
The largest TSP instance considered here contains 71,009 nodes.

\vspace{+0.5mm}\noindent\textbf{Competitors and Hyperparameter Settings.}
UNiCS was compared against state-of-the-art heuristic methods EAX~\cite{NagataK13} and LKH version 3~\cite{LKH3:helsgaun2017extension}, as well as hybrid solvers NLKH~\cite{Neurolkh:conf/nips/XinSCZ21}, VSR-LKH~\cite{VSR-LKH:journals/kbs/ZhengHZJL23}, and RHGA~\cite{RHGA:journals/cor/ZhengZCH23}.
To verify the effectiveness of UNG in guiding $AB$-cycle selection, a version of UNiCS (named UNiCS-P) that removes LS phase and uses only PBS phase was also included in the comparison.
Essentially, UNiCS-P is a variant of EAX equipped with UNG guidance. 
For all EAX-based methods (including EAX and UNiCS-P), the EAX-specific hyperparameters were set according to the original EAX paper.
All LKH-based methods (including LKH, NLKH, and VSR-LKH) utilized LKH's default hyperparameter settings.
NLKH and UNiCS shared the same SGN architecture and training procedure.
For RHGA, hyperparameters were set as specified in the original paper.
The probability of random selection of $AB$-cycles in UNiCS, i.e., $\eta$, was set to 0.5 based on the preliminary experiments. 

% Additionally, we compared our approach with LEHD\cite{LEHD}, one of the state-of-the-art deep learning-based method. However, due to its methodological limitations on large-scale TSP instances derived from real-world scenarios, LEHD significantly underperformed compared to our approach, with a relative gap spanning multiple orders of magnitude observed under same runtime budgets. We therefore present this comparison in the Appendix~E.

Additionally, a comparison was conducted with LEHD\cite{LEHD} , a state-of-the-art deep learning-based method for TSP. 
Under same runtime on Medium Set(18 instances, 5,000-10,000 nodes) representative of real-world logistics scenarios, LEHD exhibited a 62.7\% gap from BKS, exceeding UNiCS's performance (0.07\%) by over three orders of magnitude. The relative performance gap between LEHD and UNiCS widens by 3.06× when scaling from 5,000 to 10,000 nodes, revealing fundamental limitations in neural architectures for real-world TSP optimization. As these medium-sized instances remain smaller than the instance in Large set, the full detail of this comparison is relegated to Section~E in the supplementary material\cite{liu2025cascaded}.

\vspace{+0.5mm}\noindent\textbf{Testing Methods.}
To make fair comparison, the testing of all methods was conducted on the same server with two AMD EPYC 7713 CPUs, a NVIDIA A800 80G and 512GB RAM running Ubuntu 20.04, and each method was evaluated using a single CPU core with 10 independent runs per instance.
To ensure reproducibility, random seeds started from 42 and were incremented by 60 for each subsequent run.
Three runtime budgets were considered in the experiments: short (1800s), medium (3600s), and long (7200s).
All reported gaps in the experiments represent the relative gap between the algorithm's results and the Best-Known Solution (BKS),  where BKS refers to the optimal solution computed by exact algorithms or the best solution currently documented in the scientific literature for specific TSP instances.
Due to significantly inferior performance compared to other methods, the RHGA results are omitted in the paper and presented in Section~A.2 in the supplementary material\cite{liu2025cascaded}.

% The appendix, code, datasets, and instructions for reproducing the experiments are in the supplementary materials.

% \begin{figure*}[h]
%   \centering
%     \begin{subfigure}[ht]{0.34\textwidth}
%     \includegraphics[width=\linewidth]{Figure/Experiment2/TSPLib/SortGSTS1800_V8.pdf}
%     \caption{\tiny Runtime = 1800s.}
%     \end{subfigure}
%     \hfill
%     \begin{subfigure}[ht]{0.32\textwidth}
%     \includegraphics[width=\linewidth]{Figure/Experiment2/TSPLib/SortGSTS3600_V6.pdf}
    
%     \caption{\tiny Runtime = 3600s.}
%     \end{subfigure}
%     \begin{subfigure}[ht]{0.32\textwidth}
%     \includegraphics[width=\linewidth]{Figure/Experiment2/TSPLib/SortGSTS7200_V7.pdf}
%     \caption{\tiny Runtime = 7200s.}
%   \end{subfigure}
%   % \vspace{-2mm}
%   \caption{Cumulative gaps on Large set with instances sorted by ascending problem size.}
% % \vspace{-2mm}
%   \label{fig3}
% \end{figure*}

\subsection{Results and Analysis}
\vspace{+0.5mm}\noindent\textbf{Results on Large Set.}
The results are reported in Table~\ref{table1}, where ``Best'' and ``Avg'' represent the best and average solution quality obtained from 10 runs, repsectively.
Overall, UNiCS is the best-performing solver in Table~\ref{table1}.
On National and VLSI benchmarks, which constitute the majority of testing instances, UNiCS consistently outperforms all competitors across all runtime budgest, in terms of average solution quality.
For best solution quality, UNiCS leads in most cases except for National benchmark at 3600s. 
On TSPLib benchmark, UNiCS slightly under-performs UNiCS-P, ranking second.
Given that TSPLib comprises just 4 out of 32 instances (12.5\%), UNiCS demonstrates the strongest overall performance across the Large set.
Additionally, the results show that EAX-based methods generally perform better than LKH-based methods on large-scale TSP instances.

Figure~\ref{fig2}(a) illustrates the convergence curves (in terms of optimality gaps) on the Large set.
It can be observed that UNiCS exhibits faster convergence in early stages compared to EAX-based methods, while maintaining better late-stage exploration capabilities than LKH-based methods, verifying that UNiCS effectively combines the strengths of both LS and PBS.
UNiCS-P, in comparison to EAX, exhibits faster early-stage convergence without compromising late-stage exploration capabilities.
To provide a more detailed analysis of performance on individual instances within the Large set, cumulative optimality gaps were calculated, where $C_{gap}(j) = \sum\nolimits_{i=1}^j gap_j$ and $gap_j$ represents the optimality gap on the $j$-th instance.
As shown in Figure~\ref{fig3}(a), UNiCS demonstrates an increasingly pronounced advantage over other competitors as problem size grows, maintaining its lead across all three runtime budgets.
Once again, UNiCS-P shows improvement over EAX on larger instances at 1800 seconds, which can be attributed to its faster convergence speed due to the neural-guided $AB$-cycle selection.
Finally, as expected, LKH-based methods perform comparably to EAX-based methods only on smaller instances, with their performance gap widening as problem size increases.

% \begin{figure}[h]
%   \centering
%   \begin{subfigure}[ht]{0.50\columnwidth}
%     \includegraphics[width=\linewidth]{Figure/Experiment2/TSPLib/SortGSTL3600_V7.pdf}
%     \caption{\tiny Runtime = 3600s.}
%   \end{subfigure}
%   \hfill
%   \begin{subfigure}[ht]{0.48\columnwidth}
%     \includegraphics[width=\linewidth]{Figure/Experiment2/TSPLib/SortGSTL7200_V7.pdf}
%     \caption{\tiny Runtime = 7200s.}
%   \end{subfigure}
%   \caption{Cumulative gaps on Ex-Large set with instances sorted by ascending problem size.}
%   \label{fig4}
% \end{figure}

\vspace{+0.5mm}\noindent\textbf{Results on Ex-Large Set.}
Since Ex-Large set contains only 6 instances, results on individual instances are reported in Table~\ref{table2}.
Results under the short runtime budget (1800s) are omitted as no method could consistently produce effective solutions within this budget.
Overall, UNiCS demonstrates superior performance on the Ex-Large set compared to all competing methods.
For both medium (3600s) and long (7200s) runtime budgets, UNiCS achieves better solution quality in terms of both best and average results across most instances.
An exception occurs with instance ch71009 (71,009 nodes) under the 3600s budget, where NLKH outperforms UNiCS due to its rapid convergence capabilities.
However, when the runtime extends to 7200s, UNiCS significantly outperforms NLKH on this instance.
Figure~\ref{fig2}(b) illustrates the convergence curves on the Ex-Large set.
Similar to observations on the Large set, UNiCS exhibits significantly faster convergence compared to EAX-based methods while demonstrating superior late-stage exploration capabilities compared to LKH-based methods.
Additionally, UNiCS-P shows improved convergence over EAX, benefiting from the UNG guidance.
Figure~\ref{fig4} presents the cumulative gaps on the Ex-Large set, showing that UNiCS maintains a lead over other methods on nearly every instance, with the margin of superiority increasing with problem size.

% With a runtime budget of 7200 seconds, UNiCS outperforms all competitors on all instances except for fna52057. 
% At 3600 seconds, due to the large size of ch71009 with 71009 nodes, NLKH outperforms other solvers on this instance.
% While UNiCS slightly trailed behind NLKH on ch71009, it remained significantly ahead of all other solvers while maintaining its lead on all other instances.
% In summary, within the Ex-Large benchmark, UNiCS demonstrated superior performance over other baselines on all instances except fna52057.

\begin{table*}[t]
\centering
\caption{Relative gap (\%) to the BKS on the Ex-Large set (10 runs per instance). ``-'' in Best / Avg columns indicates no solutions produced in 10 runs / solutions produced only in some runs. Bold indicates best performance. ``Total'' row summarizes average gaps across Ex-Large set.}
% \vspace{-2mm}
\label{table2}
\resizebox{0.95\textwidth}{!}{
\begin{tabular}{cccccccccccccc}
\toprule
\multirow{2}{*}{Runtime} & \multirow{2}{*}{Instance} & \multicolumn{2}{c}{NLKH} & \multicolumn{2}{c}{LKH} & \multicolumn{2}{c}{VSR-LKH} & \multicolumn{2}{c}{EAX} & \multicolumn{2}{c}{UNiCS-P} & \multicolumn{2}{c}{UNiCS} \\ \cmidrule(lr){3-4} \cmidrule(lr){5-6} \cmidrule(lr){7-8} \cmidrule(lr){9-10} \cmidrule(lr){11-12} \cmidrule(lr){13-14}
& & Best & Avg & Best & Avg & Best & Avg & Best & Avg & Best & Avg & Best & Avg \\ \midrule
\multirow{7}{*}{3600s} & rbz43748 & 0.083 & 0.114 & - & - & 0.026 & - & \textbf{0.006} & 0.007 & 0.010 & 0.014 & \textbf{0.006} & \textbf{0.007} \\
& fht47608 & 0.189 & 0.219 & - & - & 0.046 & - & 0.014 & 0.015 & 0.014 & 0.016 & \textbf{0.008} & \textbf{0.012} \\
& fna52057 & 0.183 & 0.224 & - & - & - & - & \textbf{0.008} & 0.014 & 0.011 & 0.012 & 0.010 & \textbf{0.011} \\
& bna56769 & 0.122 & 0.141 & - & - & - & - & 0.030 & 0.037 & 0.013 & \textbf{0.015} & \textbf{0.007} & 0.015 \\
& dan59296 & 0.138 & 0.155 & - & - & - & - & 0.039 & 0.050 & \textbf{0.010} & 0.034 & \textbf{0.010} & \textbf{0.014} \\
& ch71009 & \textbf{0.439} & \textbf{0.465} & - & - & - & - & 1.104 & 3.237 & 0.530 & 2.410 & 0.494 & 0.523 \\
& Total & 0.192 & 0.220 & - & - & - & - & 0.201 & 0.560 & 0.098 & 0.417 & \textbf{0.089} & \textbf{0.097} \\ 
\midrule
\multirow{7}{*}{7200s} & rbz43748 & 0.068 & 0.096 & 0.018 & 0.030 & 0.022 & 0.039 & \textbf{0.006} & 0.006 & 0.006 & 0.007 & \textbf{0.006} & \textbf{0.006} \\
& fht47608 & 0.139 & 0.172 & 0.038 & 0.058 & 0.034 & 0.057 & 0.010 & 0.013 & \textbf{0.008} & 0.014 & \textbf{0.008} & \textbf{0.010} \\
& fna52057 & 0.131 & 0.183 & 0.051 & - & 0.033 & 0.042 & 0.008 & \textbf{0.009} & 0.008 & 0.010 & \textbf{0.007} & 0.011 \\
& bna56769 & 0.100 & 0.118 & - & - & 0.034 & 0.066 & 0.013 & 0.014 & 0.013  & 0.013 & \textbf{0.007} & \textbf{0.010} \\
& dan59296 & 0.112 & 0.131 & - & - & 0.056 & - & 0.010 & 0.011 & \textbf{0.008} & 0.014 & \textbf{0.008} & \textbf{0.011} \\
& ch71009 & 0.293 & 0.329 & - & - & - & - & 0.184 & 0.728 & 0.075 & 0.312 & \textbf{0.060} & \textbf{0.079} \\
& Total & 0.141 & 0.172 & - & - & - & - & 0.039 & 0.130 & 0.020 & 0.062 & \textbf{0.016} & \textbf{0.021} \\
\bottomrule
\end{tabular}
}
\end{table*}

\begin{figure*}[tb]
\centering
\includegraphics[width=1.0\textwidth]{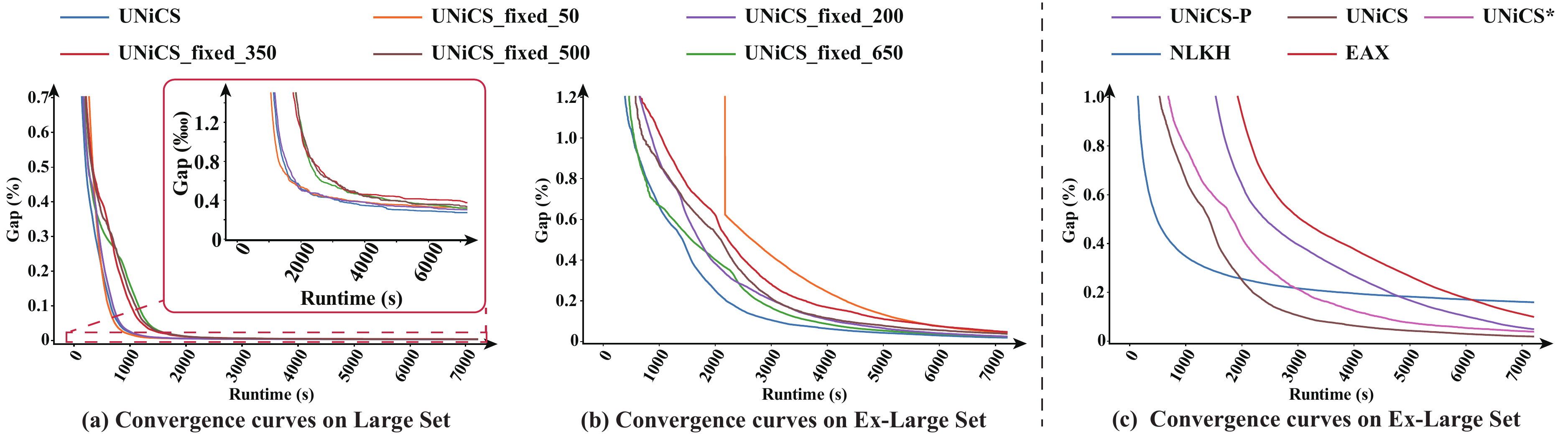}
\vspace{-3mm}
\caption{Convergence curves averaged over 10 runs in the ablation study.}
\label{fig5}
\end{figure*}

In summary, the results on both Large and Ex-Large sets affirmatively answer RQ1 raised at the beginning of this section.
That is, UNiCS generalizes effectively to these challenging TSP benchmarks and achieves superior solution quality under different runtime budgets compared to existing state-of-the-art methods.

\subsection{Ablation Study}
\noindent\textbf{Phase Transition Policy.} 
The effectiveness of the learning-based transition policy within UNG was evaluated by comparing it against fixed transition times.
During policy training, $t_{trans}$ varied from 50s to 650s. In this study, fixed $t_{trans}$ values of 50s, 200s, 350s, 500s, and 650s were tested while maintaining all other UNiCS settings.
Figure~\ref{fig5}(a) and Figure~\ref{fig5}(b) illustrate the resultant convergence curves on Large and Ex-large sets.
It can be observed that when $t_{trans}$ is fixed at a small value (e.g., 50s), UNiCS's performance on the Large set closely matches that of UNiCS using the transition policy.
However, as $t_{trans}$ increases, the performance with fixed transition times noticeably deteriorates.
Conversely, Figure~\ref{fig5}(b) demonstrates the opposite pattern: with large fixed $t_{trans}$ values (e.g., 650s), performance on the Ex-large set matches that of the transition policy, but deteriorates significantly as $t_{trans}$ decreases.
These results indicate that fixed $t_{trans}$ fails to adapt to instances of varying sizes, while the transition policy maintains strong performance across different instances.
A scatter plot of the collected data for training the transition policy is provided in Section~B in \cite{liu2025cascaded}, from which an approximately linear relationship between problem size and optimal transition time can be observed.

\vspace{+0.5mm}\noindent\textbf{Neural Guidance for $AB$-cycle Selection.}
Previous results on both Large and Ex-large sets demonstrates that UNiCS-P typically achieves better performance than the original EAX method.
To further validate the effectiveness of UNG guidance for $AB$-cycle selection within UNiCS's cascaded framework, an ablation study was conducted by replacing PBS phase in UNiCS with the original EAX method.
The resultant method is referred to as UNiCS*.
Figure~\ref{fig5}(c) shows convergence curves on Ex-large set.
The results show that UNiCS* converges significantly slower than UNiCS, confirming the effectiveness of neural guidance for $AB$-cycle selection within the cascaded framework.
Meanwhile, the fact that UNiCS* outperforms both EAX and UNiCS-P demonstrates the inherent benefits of combining the strengths of LS and PBS.

In summary, the ablation studies affirmatively answer RQ2, i.e., both the AB-cycle selection and phase transition guided by UNG contribute meaningfully to UNiCS’s performance.

\section{Conclusion}
% This work introduces UNiCS, which employs a UNG module to direct both LS and PBS, effectively combining their complementary strengths for solving large-scale TSP instances.
% With UNG guiding solution generation and determining phase transition timing, UNiCS demonstrates superior solution quality compared to state-of-the-art heuristic and hybrid methods across various runtime budgets.
% The strong performance on challenging benchmarks, despite being trained only on simple distributions, demonstrates UNiCS's generalization capabilities. Ablation studies further validate the effectiveness of UNG in guiding AB-cycle selection and phase transition.
This work introduces UNiCS, a solver that employs a Unified Neural Guidance (UNG) module to effectively combine the complementary strengths of Local Search (LS) and Population-Based Search (PBS). By guiding solution generation and phase transition timing, UNiCS consistently outperforms state-of-the-art methods on large-scale TSP instances across various runtime budgets. Strong performance on challenging benchmarks demonstrates its generalization capabilities, while ablation studies further validate the effectiveness of UNG in guiding AB-cycle selection and phase transition.

% While the transition policy in UNiCS demonstrates simplicity and effectiveness, enabling competitive performance on large-scale TSP instances, opportunities remain for enhancing its adaptability. One limitation of the present work is that, while problem size is a crucial factor in identifying appropriate transition time, it may not fully determine the optimal transition time for instances beyond those tested in this work. Future research will iexplore incorporating additional instance characteristics, such as instance size, node distribution, and hardware configuration, to develop more adaptive transition policy, thereby enhancing UNiCS’s instance- and hardware-agnostic generalization capabilities.\
While the transition policy in UNiCS is simple and effective, its adaptability can be further improved. A key limitation is its reliance on problem size, which may not be optimal for all instances. Future research will focus on developing a more adaptive policy by incorporating additional instance characteristics (e.g., node distribution) and hardware configurations, thereby enhancing UNiCS's instance and hardware-agnostic generalization. Additionally, with the rise of Large Language Models (LLMs), recent work has demonstrated their potential for guiding optimization solvers\cite{liu2024large}, which also represents a promising direction for our future research.

\begin{ack}
This work was supported by National Key Research and Development Program of China under Grant 2022YFA1004102, in part by Zhongguancun Academy Project No.20240303, in part by National Natural Science Foundation of China under Grant 72401105, and in part by Hubei Provincial Natural Science Foundation of China under Grant 2024AFB338.
\end{ack}

% %%%%%%%%%%%%%%%%%%%%%%%%%%%%%%%%%%%%%%%%%%%%%%%%%%%%%%%%%%%%%%%%%%%%%%%%

% %%% Use this command to include your bibliography file.

\bibliography{mybibfile}
\clearpage

\appendix

\section{Complete Results}
\subsection{TSP Benchmarks and Compared Methods}
\subsubsection{TSP Benchmarks}
\begin{itemize}
\item \textbf{TSPLIB} \footnote{\url{http://comopt.ifi.uni-heidelberg.de/software/TSPLIB95}}: TSPLIB is a library of instances from various sources, frequently used for testing new algorithms. It includes instances of varying sizes and complexities, providing a standard reference for comparing the performance of different TSP solvers.
\item \textbf{National} \footnote{\url{http://www.math.uwaterloo.ca/tsp/world/countries.html}}: This set consists of real-world TSP instances based on the road networks of various countries. These instances are derived from geographic data and present a challenging test bed for TSP solvers due to their large size and practical relevance.
\item \textbf{VLSI} \footnote{\url{http://www.math.uwaterloo.ca/tsp/vlsi/index.html}}: This set is tailored for very large-scale integration (VLSI) circuit design problems, where the TSP is used to optimize the routing of connections on a chip. These instances are characterized by their large size and complexity, making them a rigorous test for advanced TSP algorithms.
\end{itemize}

\begin{figure*}[tbp]
  \centering
  % 第一张子图
  \subfigure[Overall convergence curves.]{
    \includegraphics[width=0.48\textwidth]{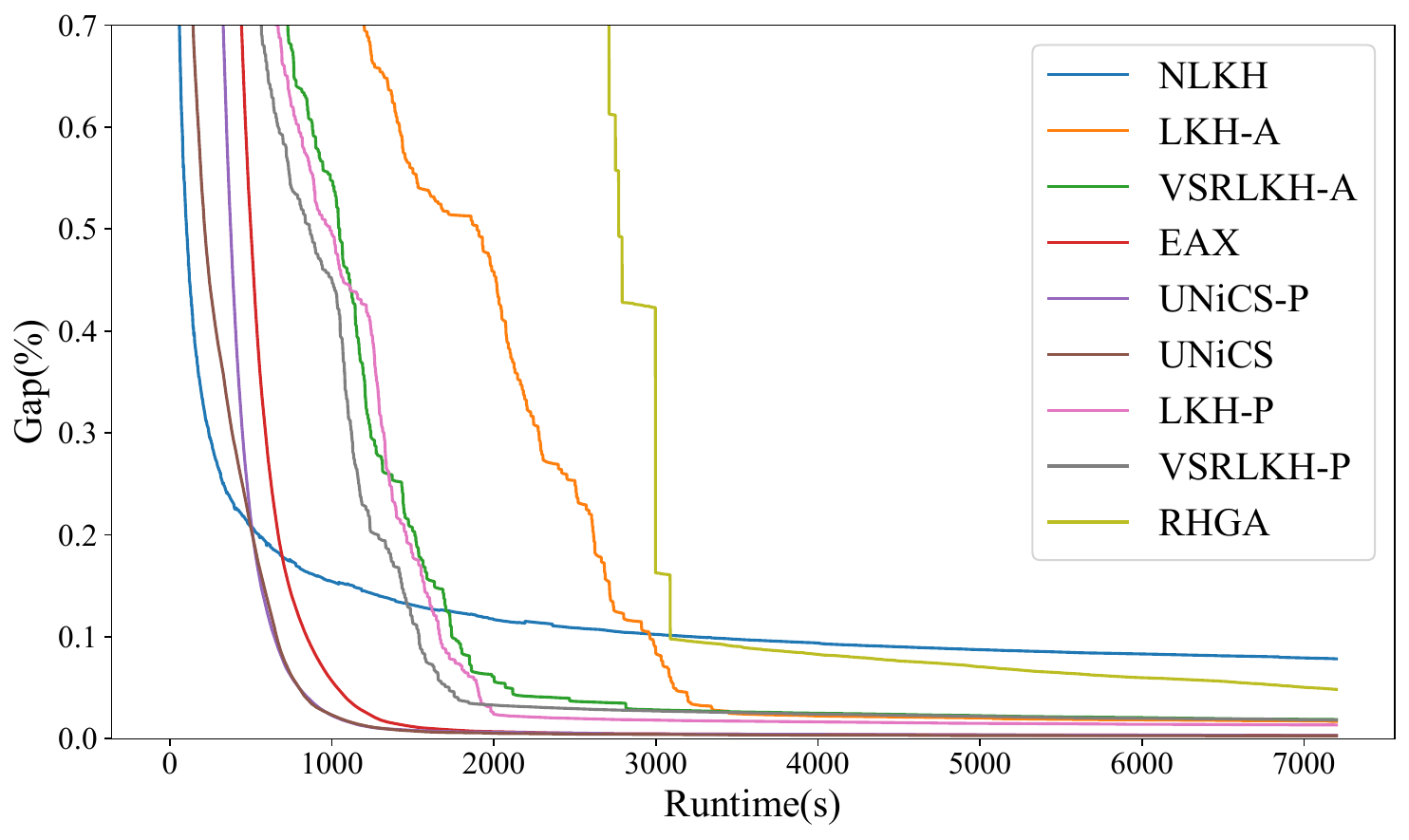}
    \label{Afig3:1}
   }
  \hfill
  \subfigure[Detailed convergence curves.]{
    \includegraphics[width=0.48\textwidth]{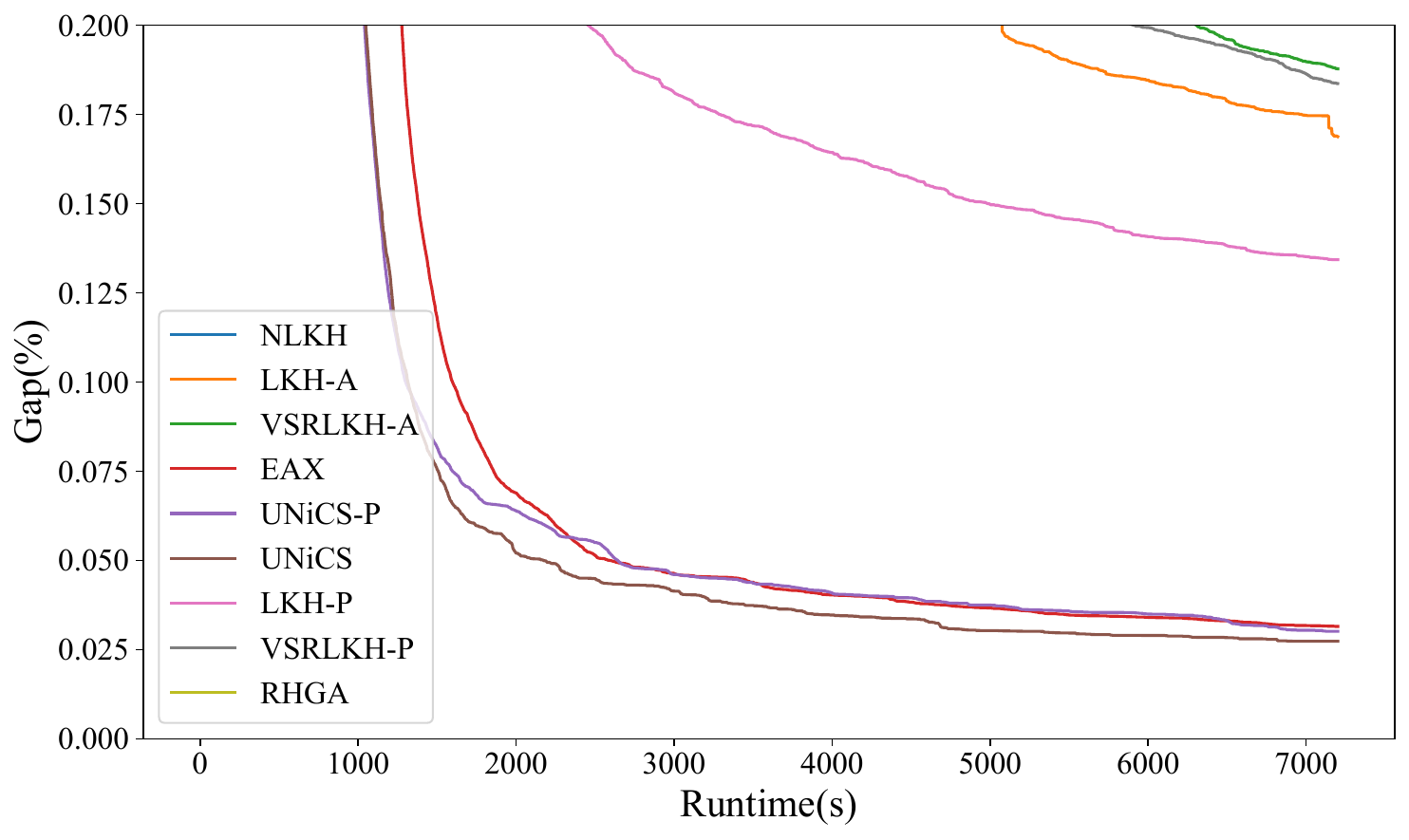}
    \label{Afig3:2}
  }
  \caption{Convergence curves in terms of optimality gaps averaged over 10 runs on Large Set.}
\end{figure*}

\begin{figure*}[tbp]
  \centering
  % 第一张子图
  \subfigure[Runtime = 1800s.]{
    \includegraphics[width=0.32\textwidth]{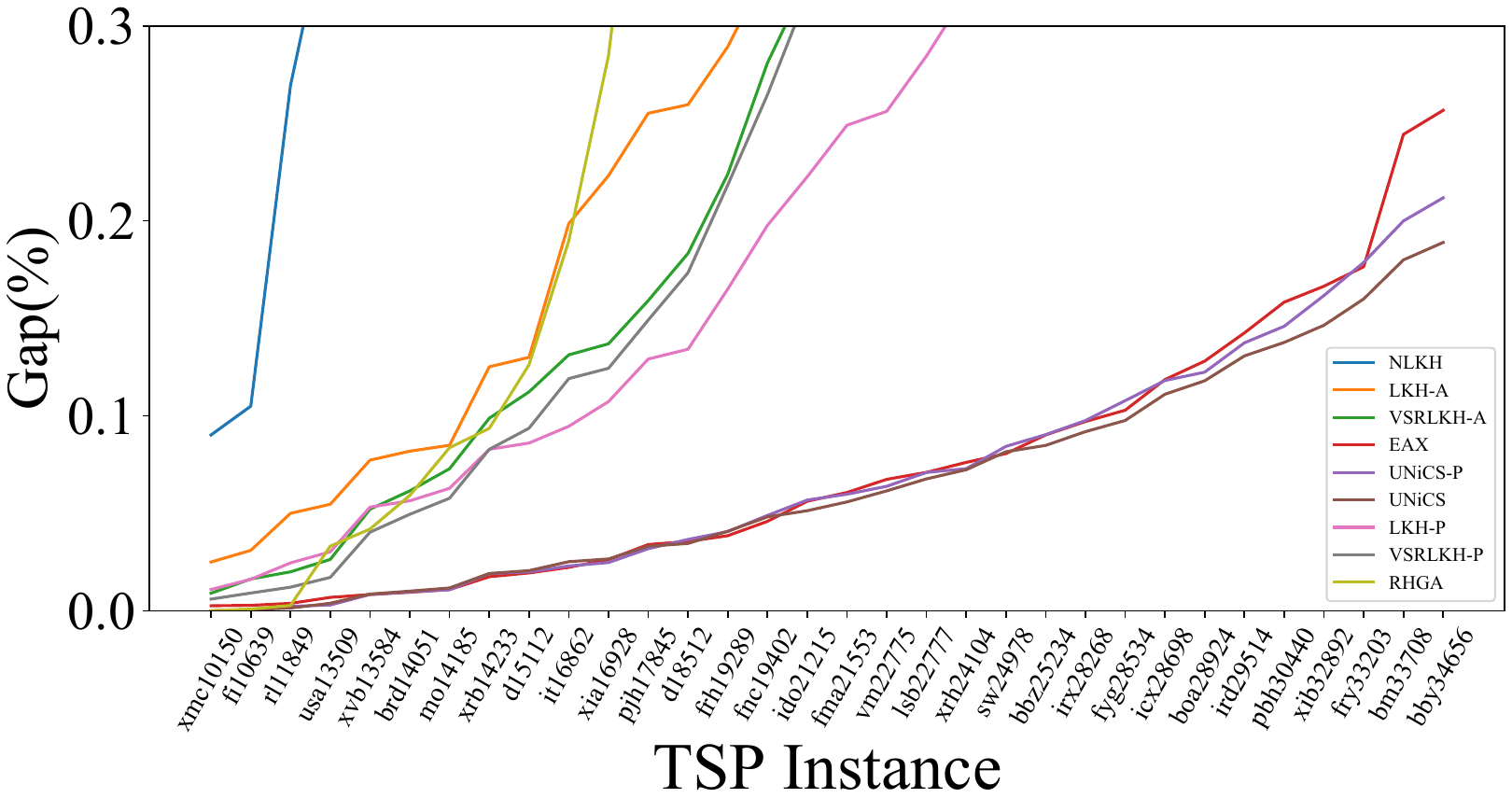}
    \label{Afig4:1}
  }
  \hfill
  % 第二张子图
  \subfigure[Runtime = 3600s.]{
    \includegraphics[width=0.32\textwidth]{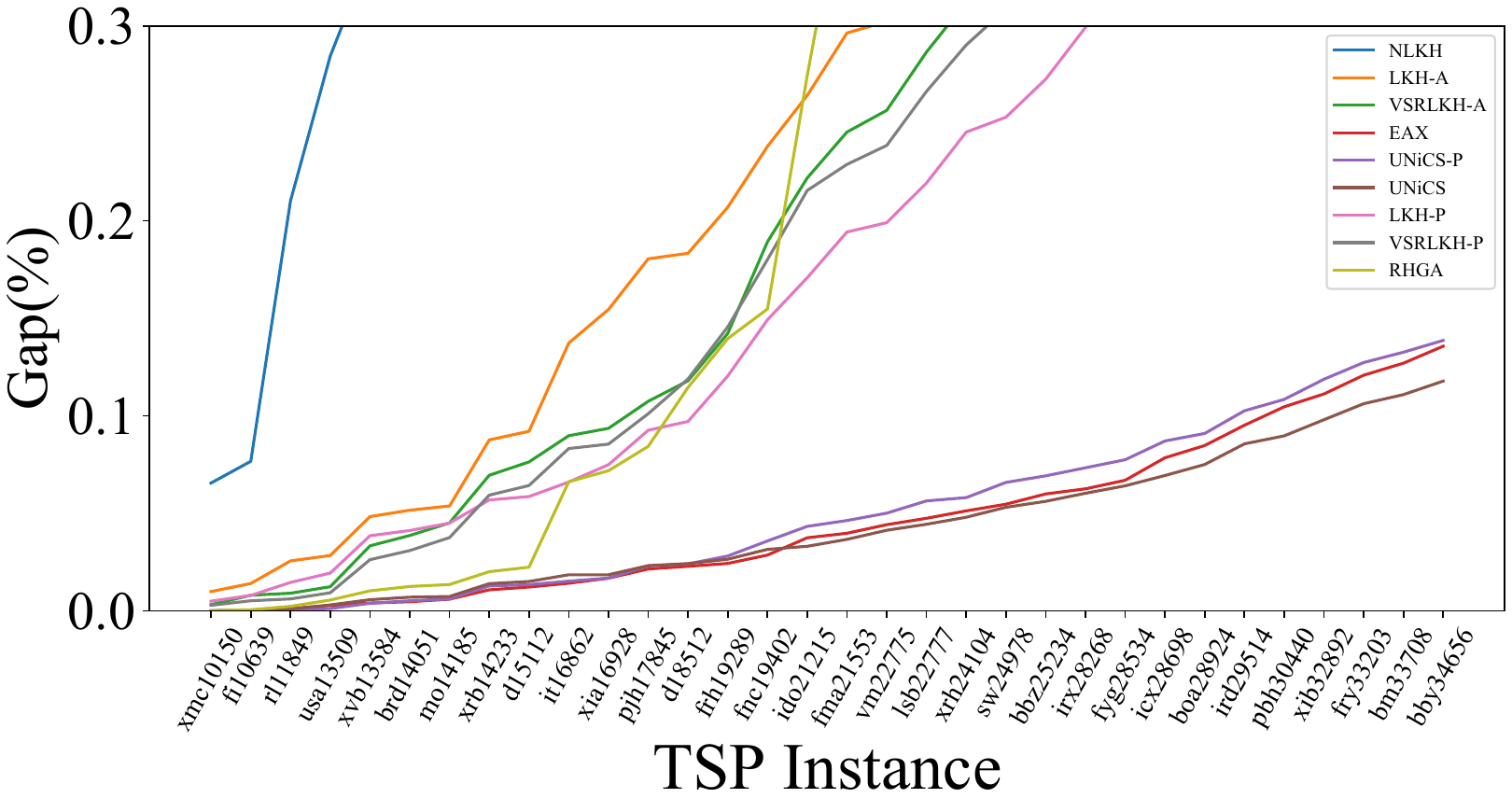}
    \label{Afig4:2}
  }
  \hfill
  % 第三张子图
  \subfigure[Runtime = 7200s.]{
    \includegraphics[width=0.32\textwidth]{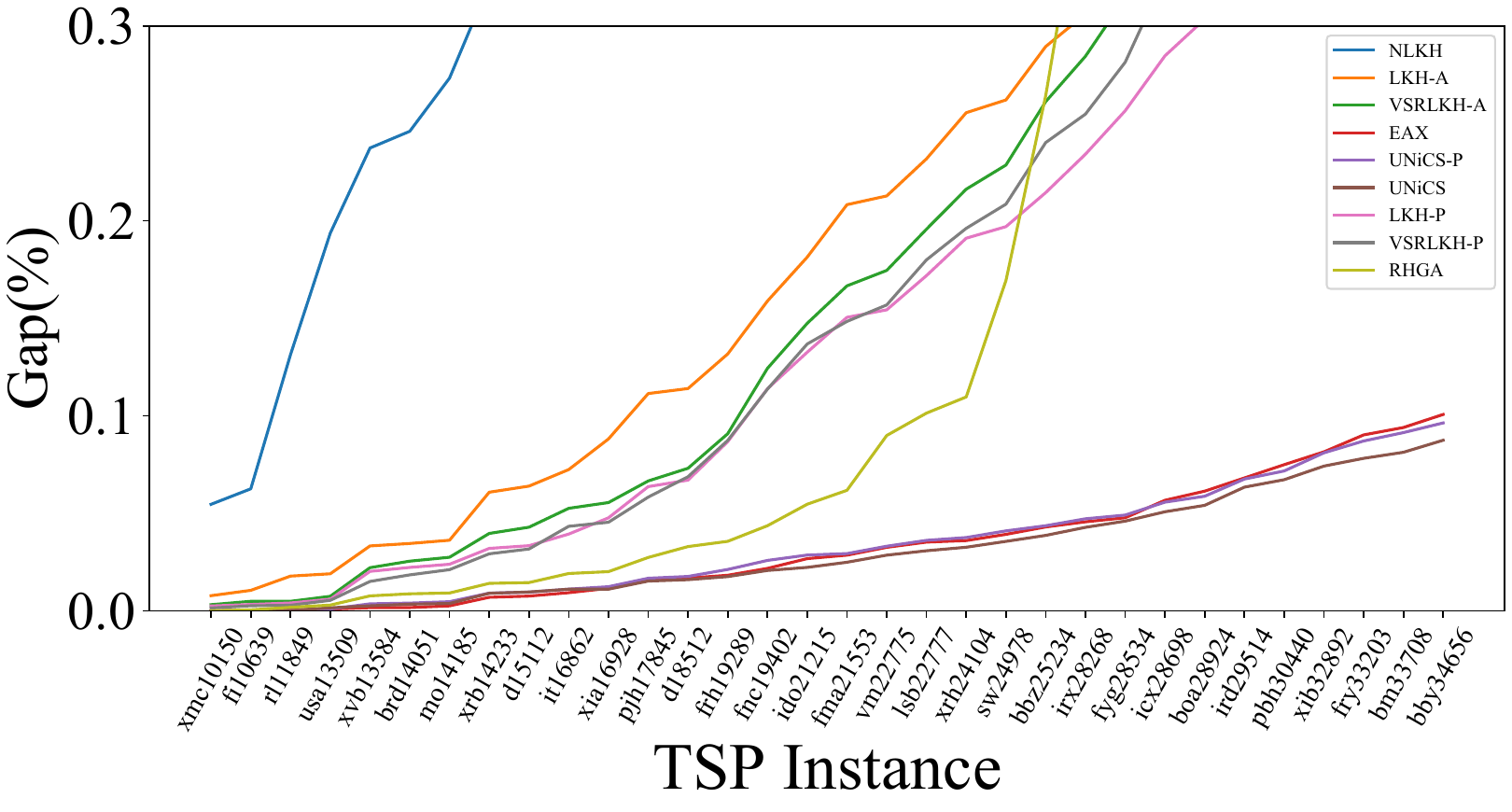}
    \label{Afig4:3}
  }
  \caption{Cumulative optimality gaps on Large set with instances sorted by ascending problem size.}
\end{figure*}

\begin{figure*}[tbp]
  \centering
  % 第一张子图
  \subfigure[Runtime = 3600s.]{
    \includegraphics[width=0.48\textwidth]{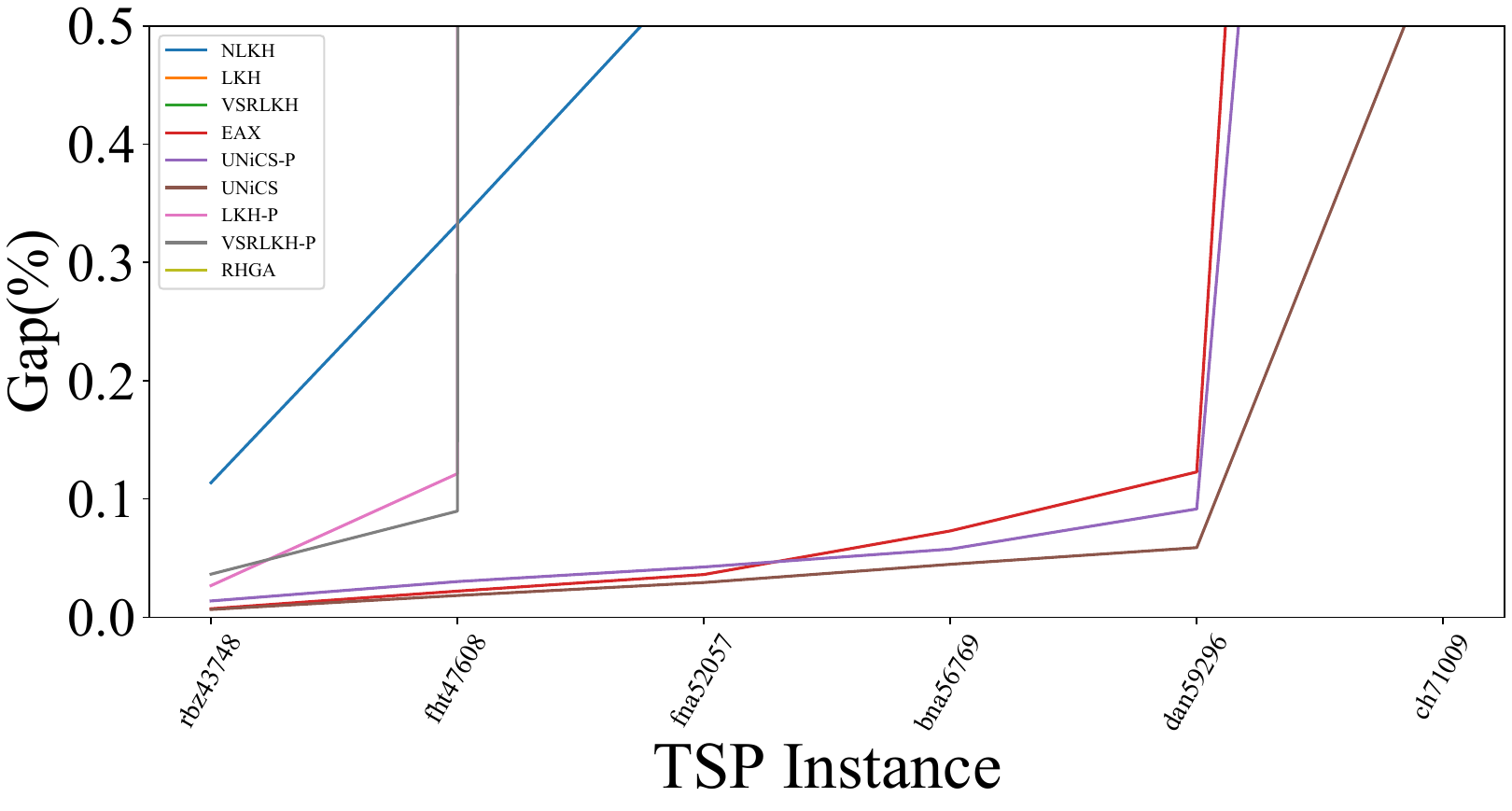}
    \label{Afig5:1}
  }
  \hfill
  % 第二张子图
  \subfigure[Runtime = 7200s.]{
    \includegraphics[width=0.48\textwidth]{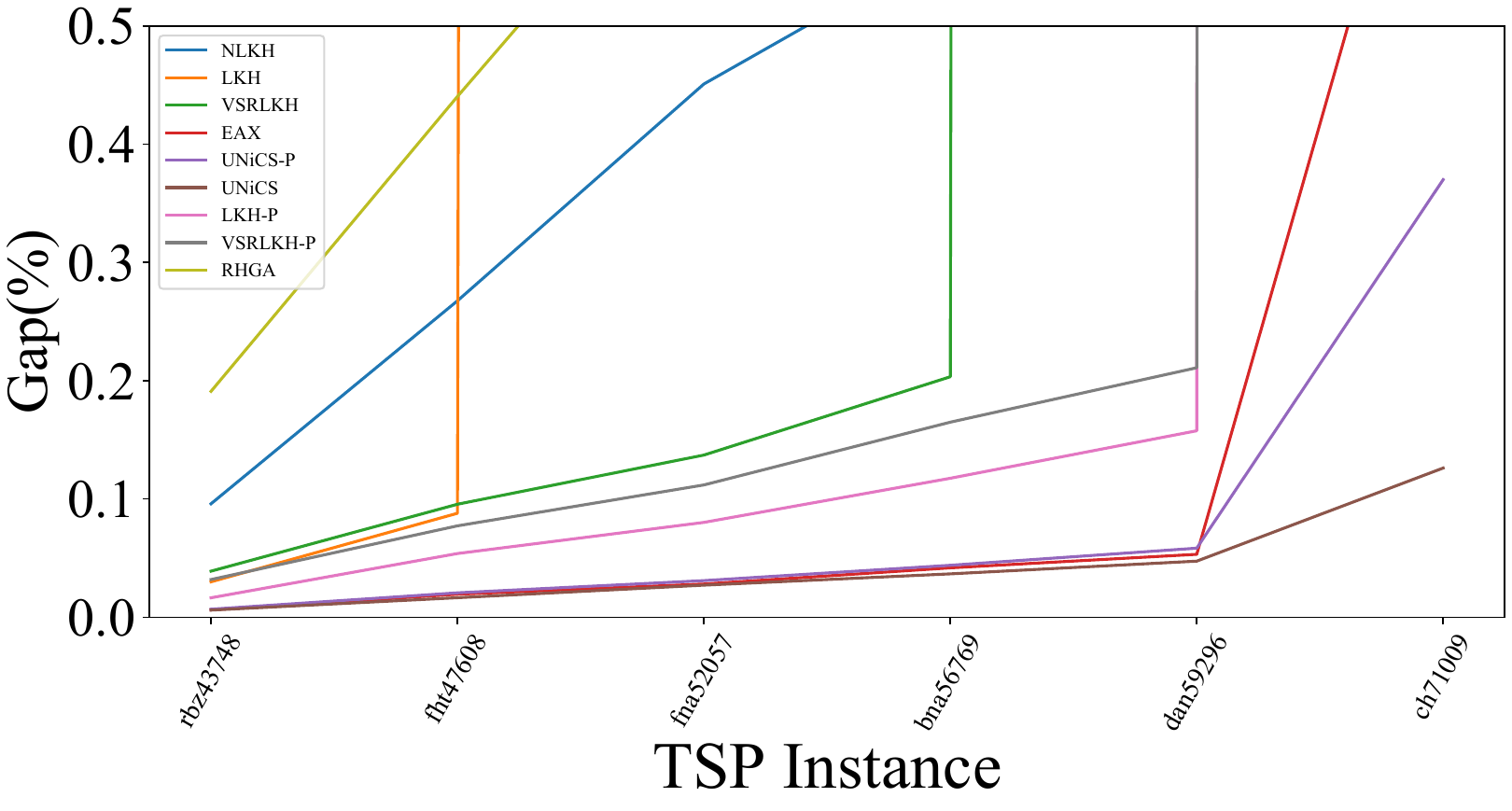}
    \label{Afig5:2}
  }
  \caption{Cumulative optimality gaps on Ex-Large set with instances sorted by ascending problem size.}  % 修正标题末尾多余的句点
\end{figure*}

\begin{figure}[!t]
\centering
\includegraphics[width=\columnwidth]{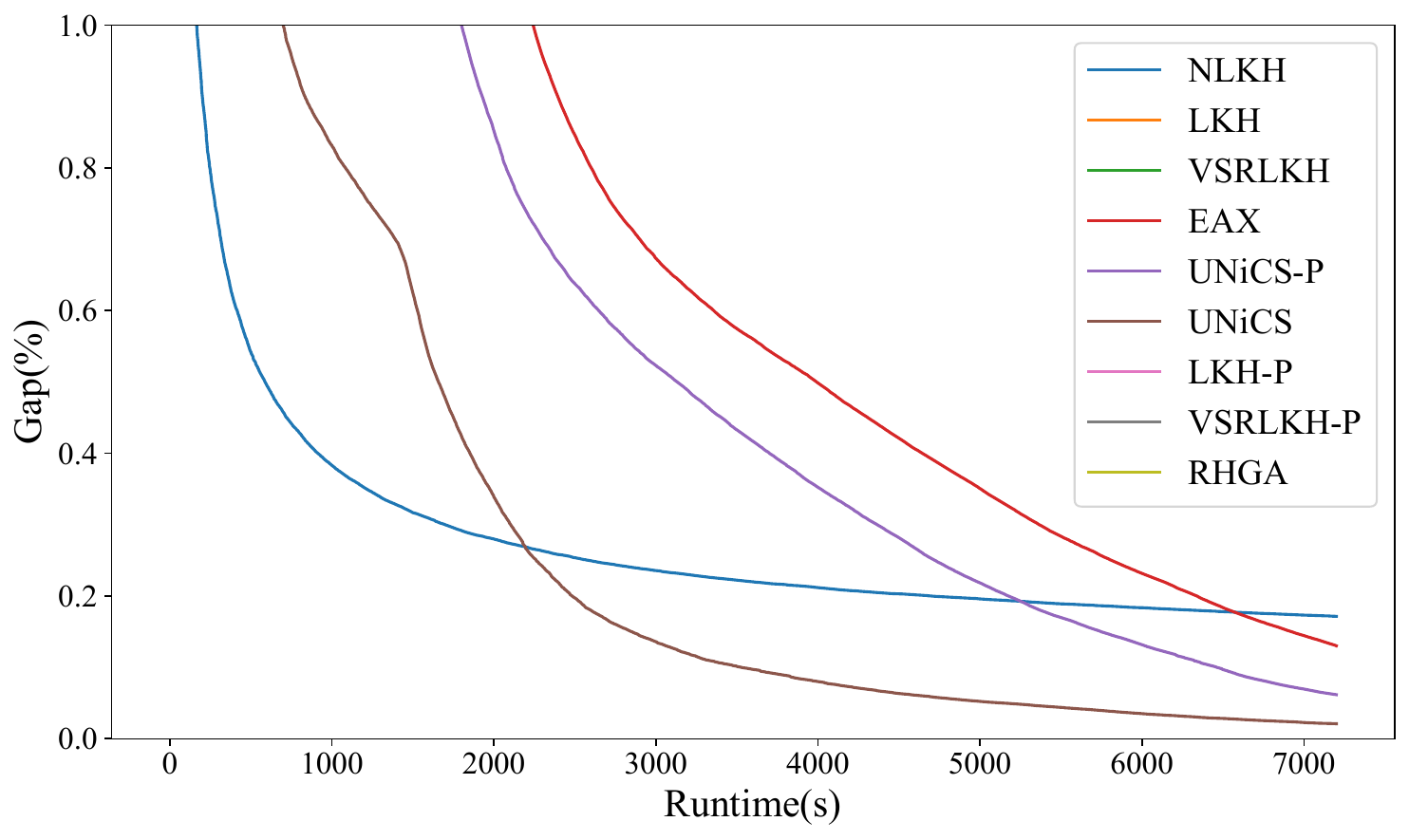} % Reduce the figure size so that it is slightly narrower than the column.
\caption{Convergence curves in terms of optimality gaps averaged over 10 runs on Ex-Large Set..}
\label{Afig6}
\end{figure}

\subsubsection{Compared Methods}
For LKH and VSR-LKH, in addition to the original implementation using $\alpha$-measure method, a variant using POPMUSIC~\cite{helsgaun2018using} was also evaluated.
\begin{itemize}
    \item \textbf{EAX}: All settings are consistent with the original EAX paper.
    \item  \textbf{LKH-A and LKH-P}: LKH-A refers to the default LKH implementation using the $\alpha$-measure method, as reported in the main text of the paper. LKH-P represents a variant of LKH using the POPMUSIC method.
    All other parameters remained consistent with the original paper.
    While LKH-P showed modest improvements over LKH-A, both versions performed significantly below UNiCS.
    \item \textbf{VSR-LKH-A and VSR-LKH-P}: Similarly, VSR-LKH-A denotes the original VSR-LKH using the $\alpha$-measure method, while VSR-LKH-P utilizes the POPMUSIC method. Despite VSR-LKH-P demonstrating some improvements over VSR-LKH-A, both versions remained substantially inferior to UNiCS in performance.
    \item \textbf{NLKH}: The problem was solved using our retrained neural network, with all other settings following the original paper.
    \item \textbf{RHGA}: Since the code for RHGA is not publicly available, we implemented the RHGA solver based on VSR-LKH.
    All the parameter settings followed the original paper.
    % with a specific exception for $M_{gen}$, which was set to a constant value of 30 due to the sample sizes tested ranging from 10,000 to 100,000.
\end{itemize}

\subsection{Results}
\subsubsection{Large Set}
Fig.\ref{Afig3:1} and Fig.\ref{Afig3:2} further illustrate the overall convergence curves of the solvers on the Large Set.
The figures show that the UNiCS consistently outperforms all other methods across the entire runtime.
It effectively combines the strengths of LS and UNiCS-P, achieving the fastest convergence in both the early and late stages, highlighting its superior exploration and exploitation capabilities.
Furthermore, a noticeable trend is observed when comparing the performance of LKH-A and VSR-LKH-A with their counterparts LKH-P and VSR-LKH-P.
The LKH-P and VSR-LKH-P methods demonstrate significantly better performance than LKH-A and VSR-LKH-A.
This improvement underscores the POPMUSIC method's effectiveness.
The performance of RHGA is inferior compared to the other solvers.
This can be due to its design, which prioritizes a thorough exploration of the solution space to identify the optimal solution.
As a result, RHGA requires a significantly longer time to fully converge and showcase its potential in finding the optimal solution.
This detailed exploration, while potentially advantageous in the very long term, hindering its performance in achieving convergence with normal runtime budgets, particularly on larger instances.

To further demonstrate the performance of each solver on individual instances within the Large Set, we employed the cumulative optimality gap to showcase their performance.
As depicted in Fig.~\ref{Afig4:1}, Fig.~\ref{Afig4:2} and Fig.~\ref{Afig4:3}, UNiCS demonstrates a more pronounced advantage over other solvers on larger instances, maintaining its lead across different runtime budgets.
Additionally, LKH-P and VSR-LKH-P outperform their counterparts, LKH-A and VSR-LKH-A. 
Interestingly, RHGA shows competitive performance on smaller instances (e.g., xmc10150), where it even outperforms LKH-based methods. However, as the problem size increases, RHGA's performance deteriorates significantly.
This further emphasizes RHGA's design focus on thorough exploration at the cost of slow convergence.

\subsubsection{Ex-Large Set}
Fig.~\ref{Afig6} presents the convergence curves of the solvers on the Ex-Large set. Similar to the main text, NLKH demonstrates an early advantage in the initial stages of solving. However, UNiCS surpasses it in overall convergence speed by effectively combining the strengths of both UNiCS-P and LS.
Fig.~\ref{Afig5:1} and Fig.~\ref{Afig5:2} present the cumulative gap for each solver.
It can be observed that the UNiCS maintains a lead over other baselines on nearly every instance, with the margin of superiority increasing as the problem size grows.
Additionally, the advantage of LKH-P and VSR-LKH-P over LKH-A and VSR-LKH-A is maintained on Ex-Large Set.

\subsubsection{Results on each TSP instance: } 
Tables~\ref{Atable1}-\ref{Atable4} (at the end of the Appendix) present the performance of all solvers on each instance at different solving times, respectively. It can be observed that the UNiCS not only achieves the best performance in terms of the overall average gap but also demonstrates superior performance on larger instances overall.

\subsection{Neural Guidance for $AB$-cycle Selection}
Fig.~\ref{Afig7:1} and Fig.~\ref{Afig7:2} show the convergence curves of UNiCS* and other solvers on Large set and Ex-Large set, respectively.
% It can be observed that the performance of UNiCS* lies between that of UNiCS and EAX.
% On the Large Set, the superior performance of the UNiCS over UNiCS* once again demonstrates the effectiveness of UNiCS-P within the UNiCS framework.

\subsection{The Hyperparameter $\eta$}
To determine the value of $\eta$, we conducted a preliminary experiment on the performance of UNiCS-P.
We randomly selected five instances as the validation set: d15112, ho14473, isb22777, usa13509, and vm22775.
We explored five different values of $\eta$: 0, 0.25, 0.5, 0.75, and 1.
For example, when $\eta$ is set to 0.25, the solver is denoted as UNiCS-P-R0.25, and similarly for other values.
Fig.~\ref{Afig8:1_A} and Fig.~\ref{Afig8:2_A} show the performance of the algorithm on the test Set with different $\eta$ values.
Overall, the larger the $\eta$, the faster the early convergence of UNiCS-P, which is attributed to the effectiveness of the neural network guidance.
However, when $\eta$ is too large (e.g., $\eta=1$), the convergence speed of UNiCS-P in the mid-to-late stages is negatively impacted, possibly due to the loss of population diversity caused by excessive reliance on the neural network.
Considering that UNiCS-P will be combined with LS in UNiCS, we do not need to overly prioritize early convergence speed.
Therefore, we ultimately selected $\eta=0.5$ to balance early and late convergence speeds.

\section{ Transition Policy}

\subsection{Training Detail}
\begin{figure}[t]
\centering
\includegraphics[width=\columnwidth]{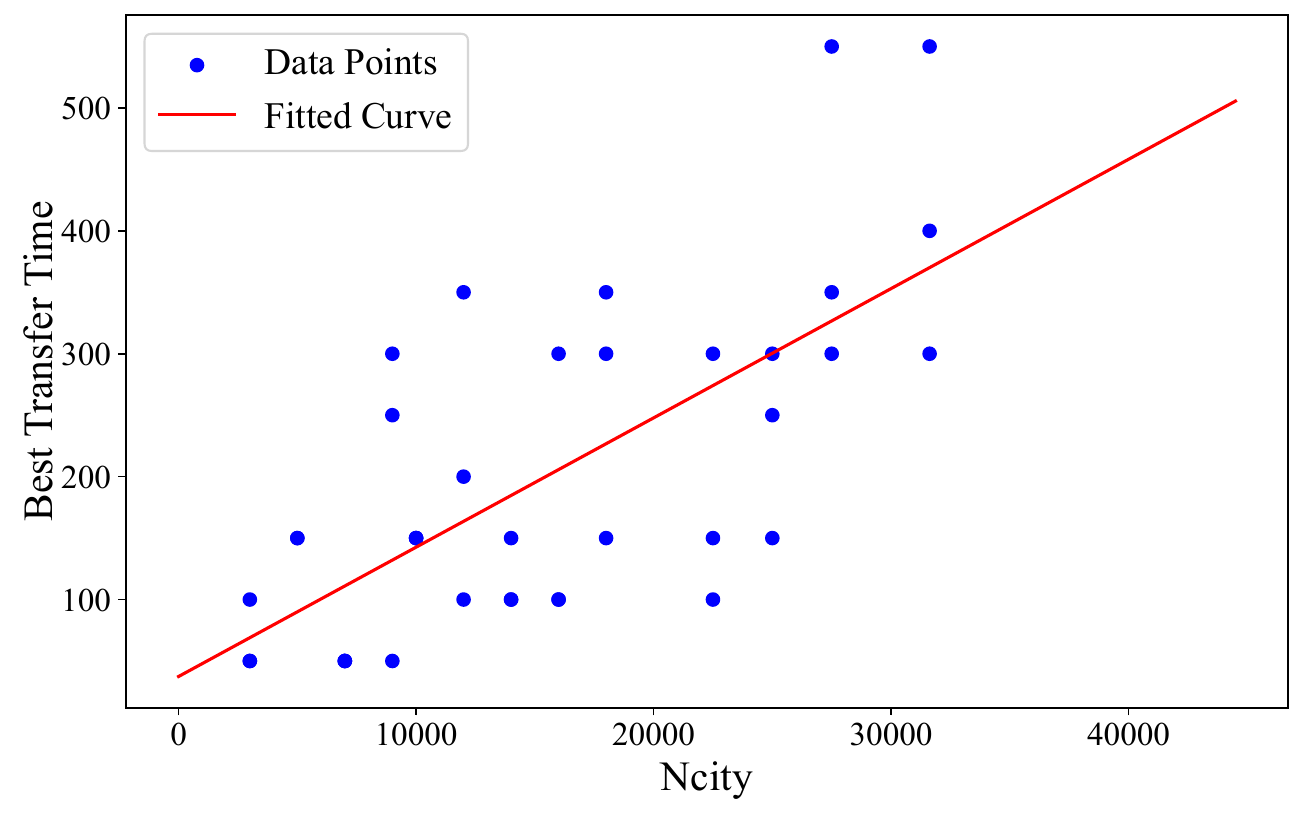} % Reduce the figure size so that it is slightly narrower than the column.
\caption{The relationship between $N_{city}$ and the optimal $t_{trans}$, where the blue data points represent the training data and the red line represents the fitted transtition policy.}
\label{Afig10}
\end{figure}

\begin{figure*}[tbp]
  \centering
  % 第一张子图
  \subfigure[Overall convergence curves.]{
    \includegraphics[width=0.48\textwidth]{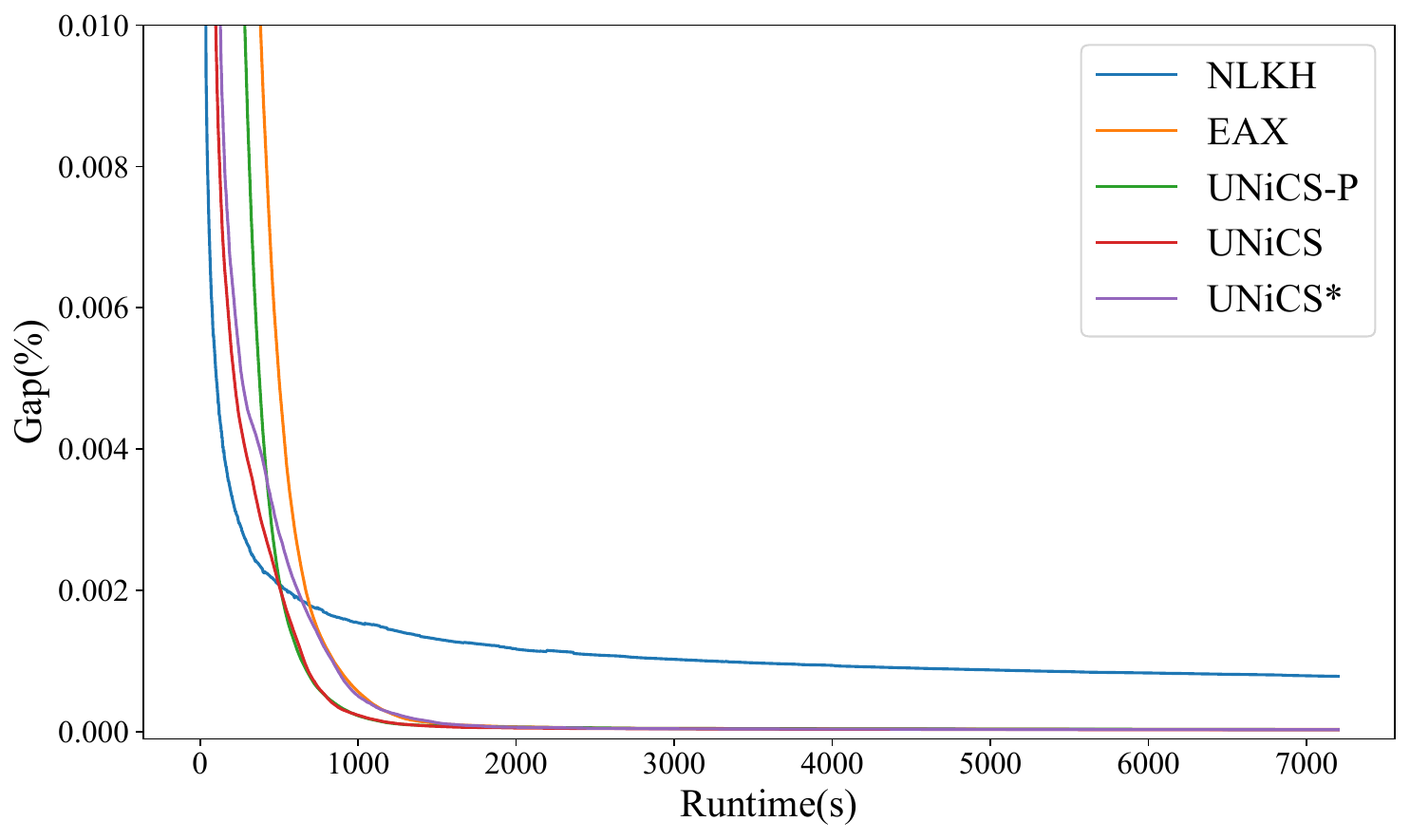}
    \label{Afig7:1}
  }
  \hfill
  % 第二张子图
  \subfigure[Detailed convergence curves.]{
    \includegraphics[width=0.48\textwidth]{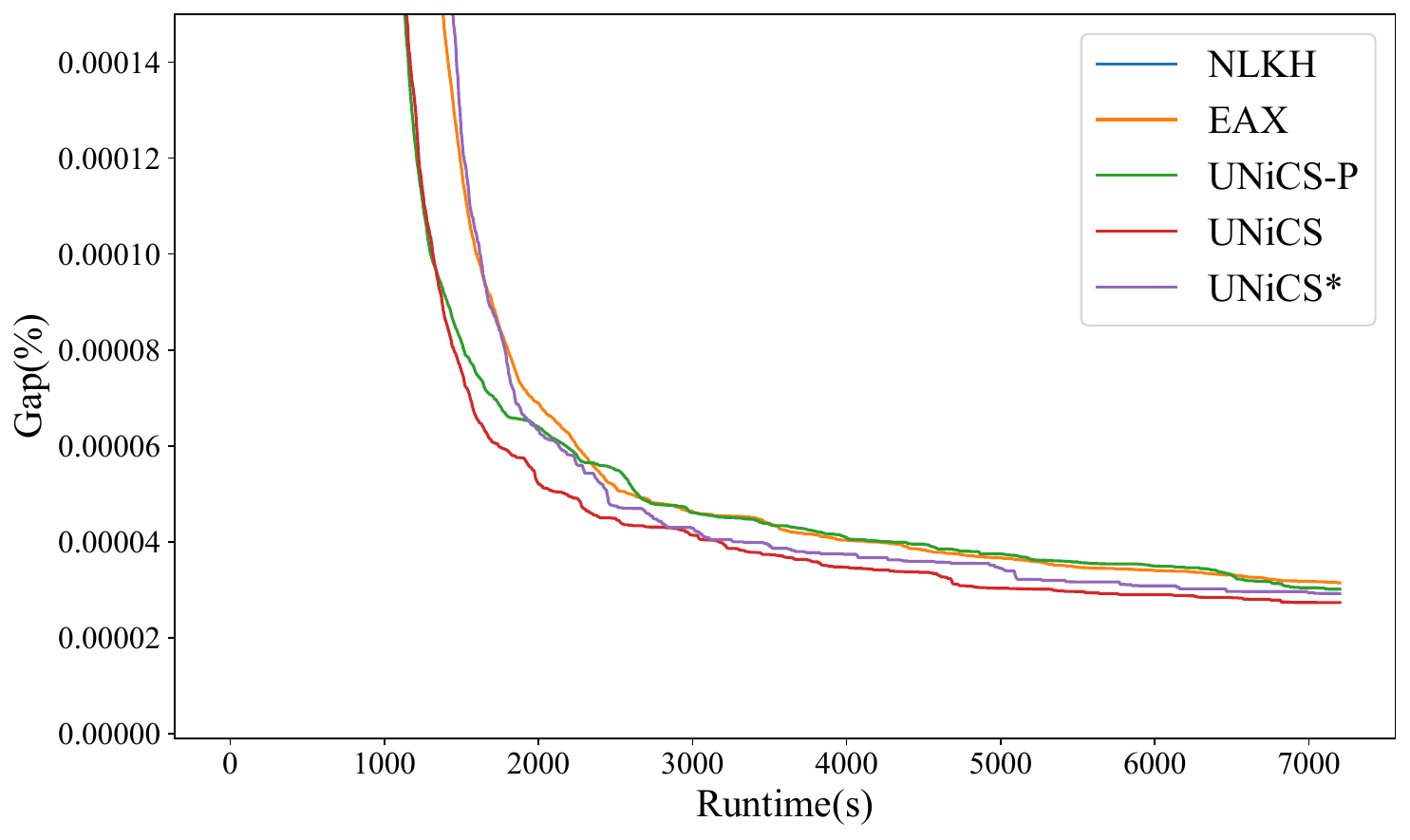}
    \label{Afig7:2}
  }
  \caption{Convergence curves in terms of optimality gaps averaged over 10 runs in the ablation study.}
\end{figure*}

\begin{figure*}[tbp]
  \centering
  % 第一张子图
  \subfigure[Overall convergence curves.]{
    \includegraphics[width=0.48\textwidth]{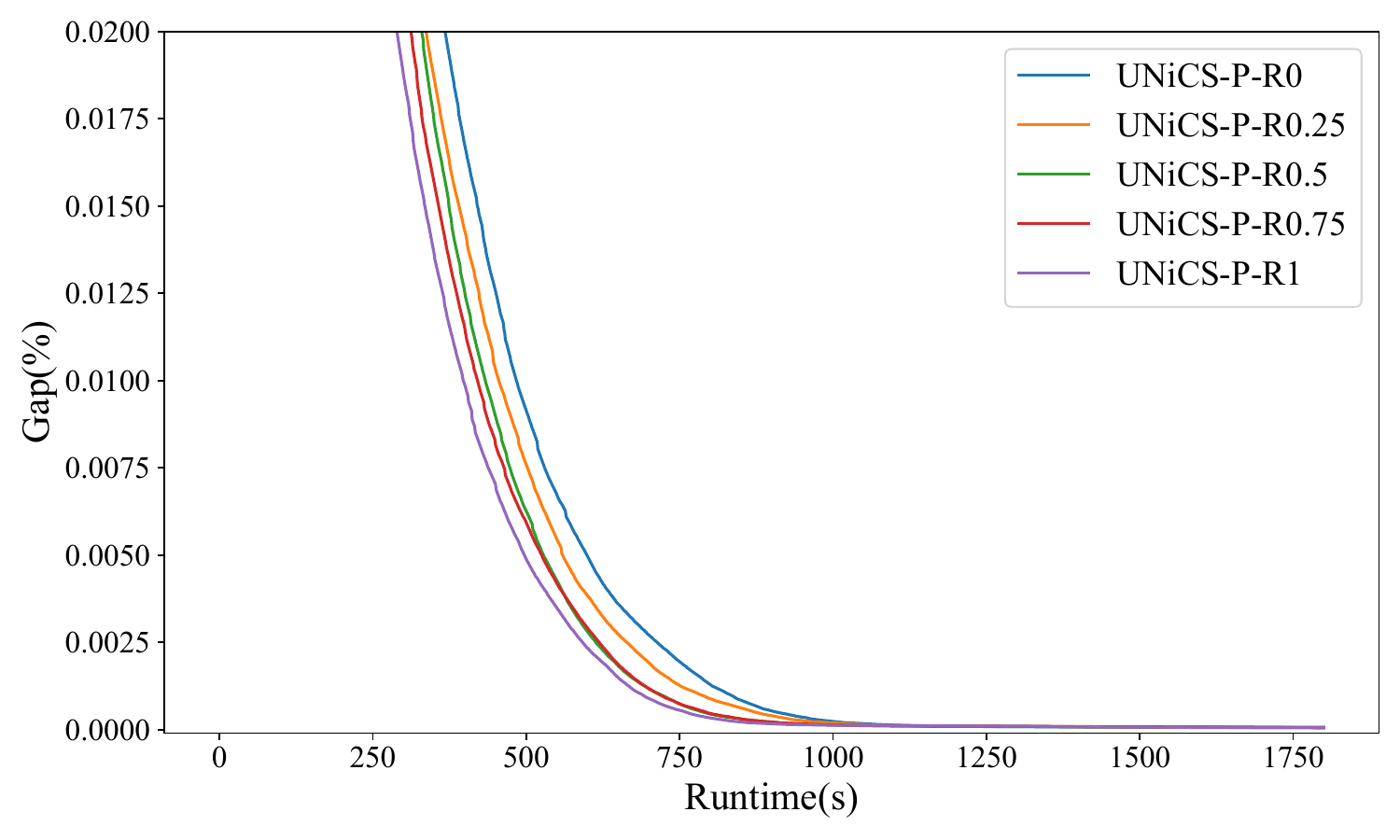}
    \label{Afig8:1_A}
  }
  \hfill
  % 第二张子图
  \subfigure[Detailed convergence curves.]{
    \includegraphics[width=0.48\textwidth]{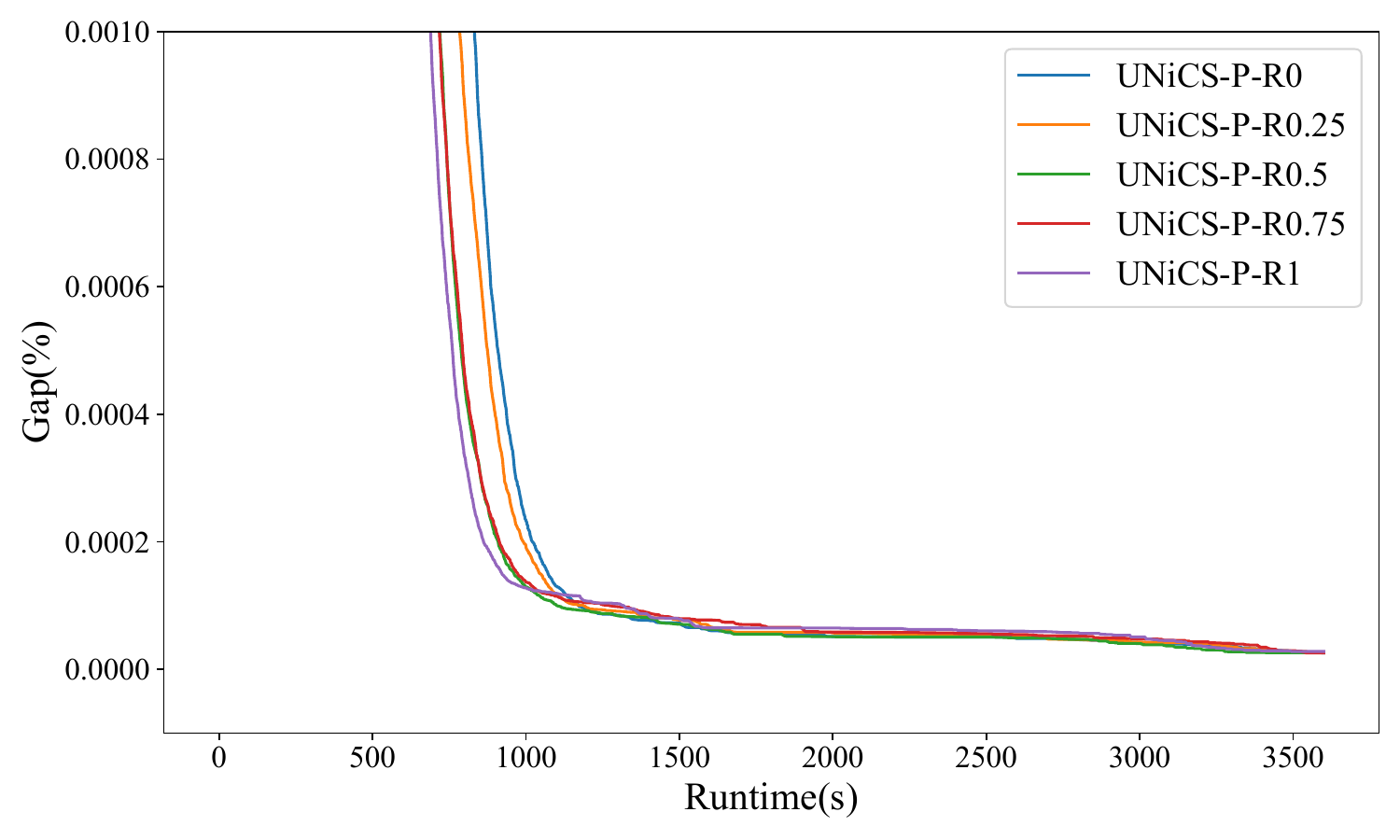}
    \label{Afig8:2_A}
  }
  \caption{Convergence curves for different values of $\eta$.}
\end{figure*}

\begin{figure*}[tbp]
  \centering
  % 第一张子图
  \subfigure[Overview]{
    \includegraphics[width=0.48\textwidth]{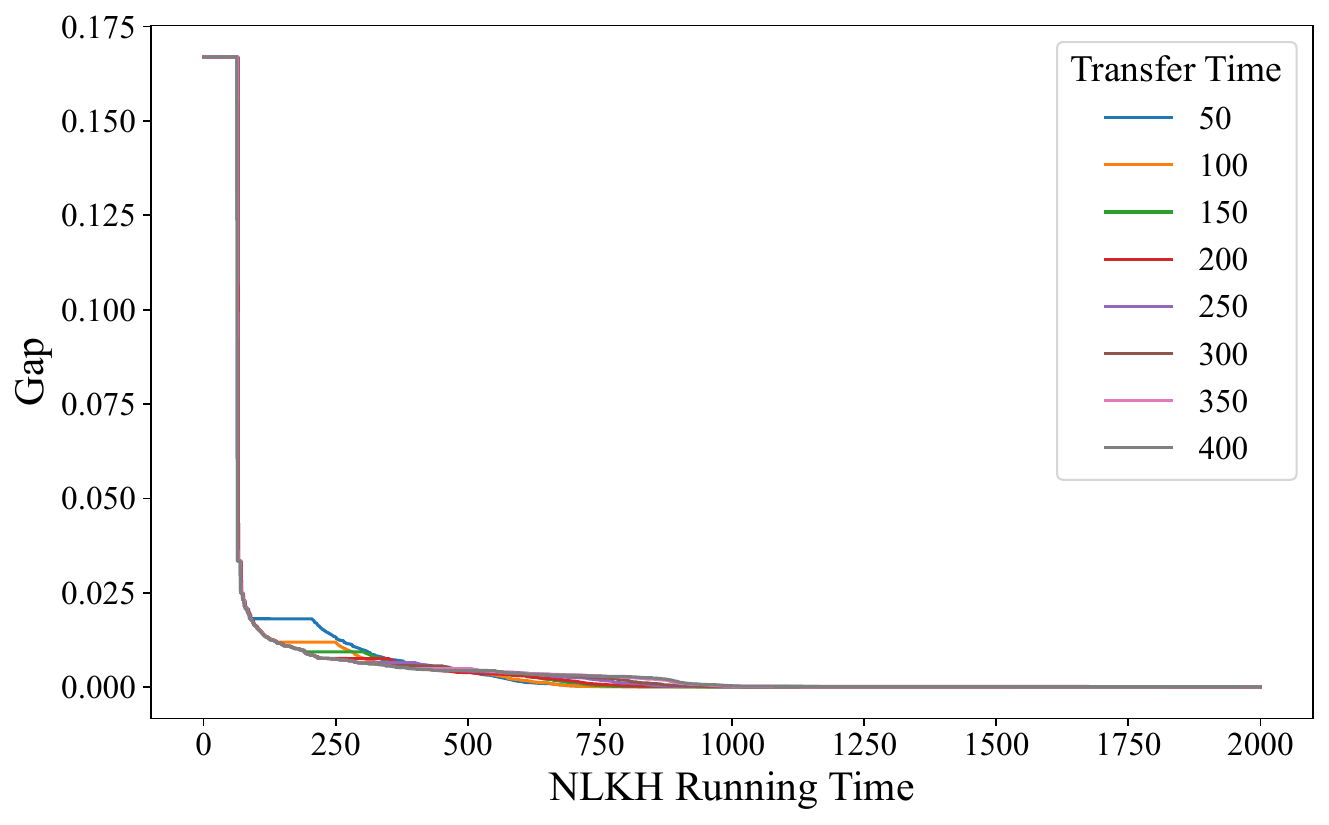}
    \label{Afig9:1}
  }
  \hfill
  % 第二张子图
  \subfigure[Detail]{
    \includegraphics[width=0.48\textwidth]{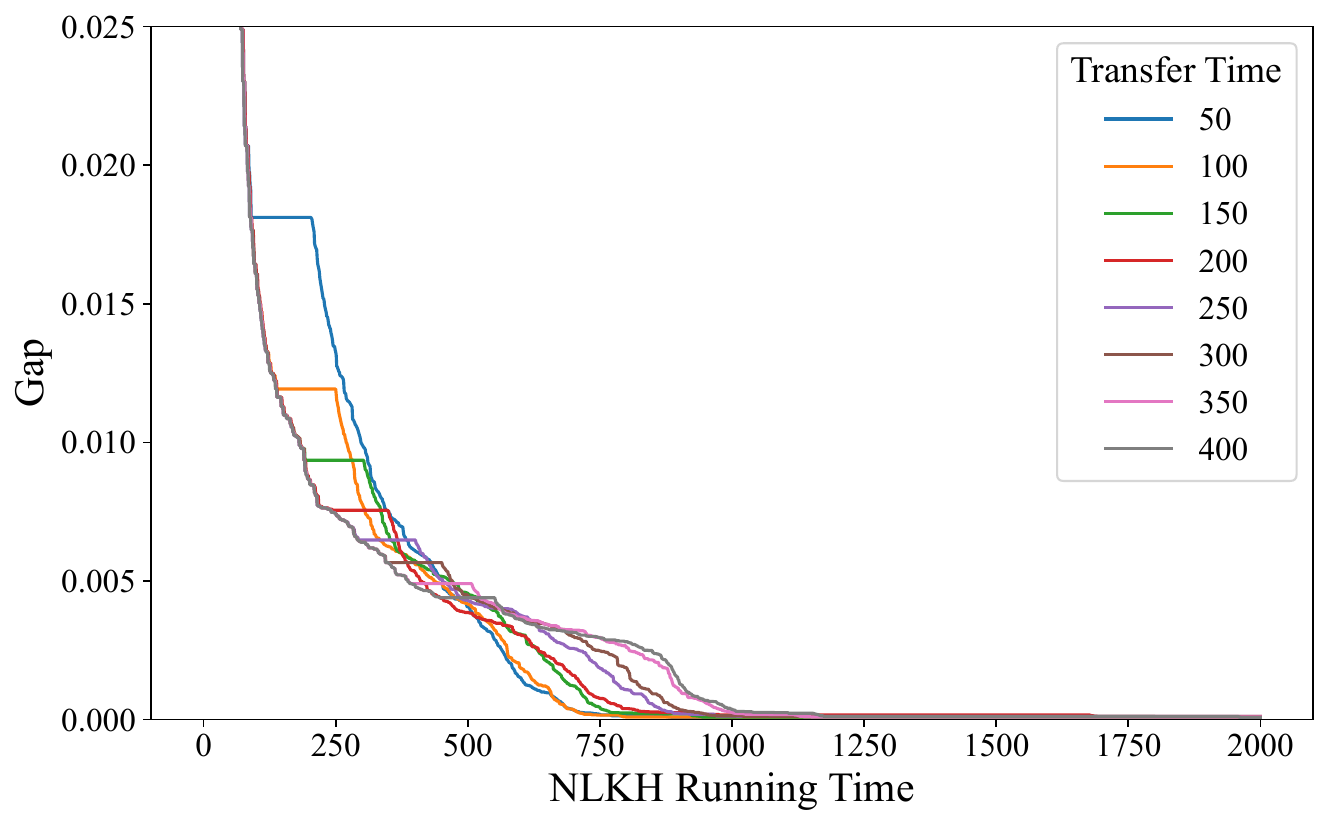}
    \label{Afig9:2}
  }
  \caption{Convergence curves for different values of $t_{\text{trans}}$ on pbh30400.}  % 修正下标格式为 \text{trans}
\end{figure*}

To train the transition policy, we selected 56 TSP instances with sizes ranging from 3,000 to 30,000, uniformly distributed as the training set. We sampled the transition time at intervals of 50 seconds, ranging from 50 seconds to 650 seconds, to obtain training data and identify the optimal $t_{trans}$ for different $N_{city}$ during training. 
Fig.~\ref{Afig10} shows the relationship between the optimal transition time and different $N_{city}$ samples in the training set, where the blue dots represent the training data.
A noticeble linear relationship between $N_{city}$ and $t_{trans}$ is observed
% Therefore, we fitted a linear model to obtain the red line in the figure, which is used as the transition Policy.
Additionally, Fig.~\ref{Afig9:1} and Fig.~\ref{Afig9:2} illustrate how $t_{trans}$ affects the convergence trend of the UNiCS on the TSP instance pbh30400.

\subsection{Evaluation Metrics}
To better train the transition strategy, we used the area under the UNiCS convergence curve as the evaluation metric. Specifically, assuming our solving time is $T$, we sample the gap at 1-second intervals, yielding a gap list of length $T$, denoted as $L_{Gap}^T$. For the area under the UNiCS convergence curve, we compute it using $Gap_{sum} = \sum\limits_{i=1}^TL_{Gap}^T(i)$, where $Gap_{sum}$ is the evaluation metric, which we refer to as the sum gap. During the entire solving process, there may be moments where no solution is available. In such cases, we record the gap as 10 times the gap corresponding to the algorithm's first valid solution as a penalty.

\section{The EAX Method}
\begin{figure}[tbp]
\centering
\includegraphics[width=\columnwidth]{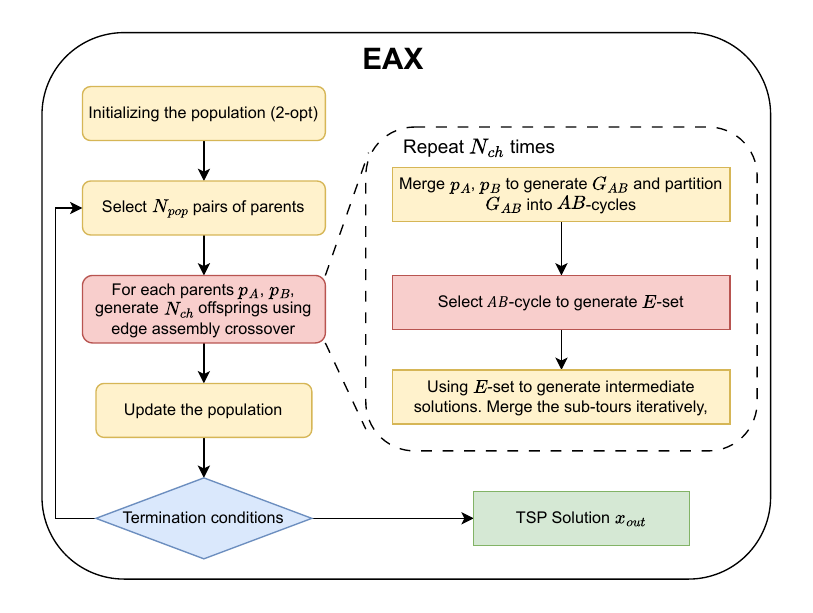} % Reduce the figure size so that it is slightly narrower than the column.
\caption{The overall structure of EAX.}
\label{EAX_Structure}
\end{figure}

The EAX method, based on a genetic algorithm, treats solutions to the TSP problem as key individuals in a population and updates the population by generating offspring through edge assembly crossover. As shown in Fig.~\ref{EAX_Structure}, the overall structure of the EAX solver is as follows: 
\begin{itemize}
    \item First, the initial population is determined using the 2-opt method.
    \item Then, the generation process is repeated according to the GA framework. In each generation of the EAX, offspring are generated from the parents using edge assembly crossover, and the population is updated based on the replacement algorithm.
    \item Finally, when the stopping criteria are met, the algorithm terminates and provides the final solution.
\end{itemize}

The assembly crossover operation is the core of the EAX method. Assuming $p_A$ and $p_B$ are the two parents used to generate offspring in one assembly crossover operation, $N_{ch}$ is the number of offspring to be generated, and $E_A$ and $E_B$ are the sets of edges included in the solutions $p_A$ and $p_B$, respectively. The EAX generates offspring through the following steps.

\begin{itemize}
\item \textbf{Step 1}: Construct the undirected multigraph $G_{AB} = (V, E_A \cup E_B)$ by taking the union of $E_A$ and $E_B$. Then, generate $AB$-cycles from $G_{AB}$ through random walks until all edges are assigned to $AB$-cycles. An $AB$-cycle is a cycle alternately formed by edges belonging to $E_A$ and $E_B$.
\item \textbf{Step 2}: Select the $AB$-cycles to construct the $E$-set according to the selection strategy of the search phase (in stage 1, Random Selection). The $E$-set is defined as the union of the selected $AB$-cycles.

\item \textbf{Step 3}: Generate an intermediate solution from $p_A$ by removing the edges in $E_A$ that belong to the $E$-set and adding the edges in $E_B$ from the $E$-set. Specifically, the intermediate solution is generated using the formula $E_C = (E_A \setminus (E\text{-set} \cap E_A)) \cup (E\text{-set} \cap E_B)$. This intermediate solution usually consists of one or more sub-tours and is therefore not typically a valid TSP solution. The smallest sub-tour (the sub-tour with the fewest edges) is then merged with other sub-tours to generate a valid offspring through a predefined merging method.

\item \textbf{Step 4}: Repeat the selection of $AB$-cycles and the generation of intermediate solutions until $N_{ch}$ offspring are created. Different offspring can be generated each time due to the varying selection of $AB$-cycles.

\end{itemize}

The algorithm consists of two phases. It terminates the first stage and switches to the second stage when no improvement in the best solution is found within a period of generations, thereby adapting to the varying difficulty levels of the search phases.

\begin{figure*}[tbp]
\centering
\includegraphics[width=1.0\textwidth]{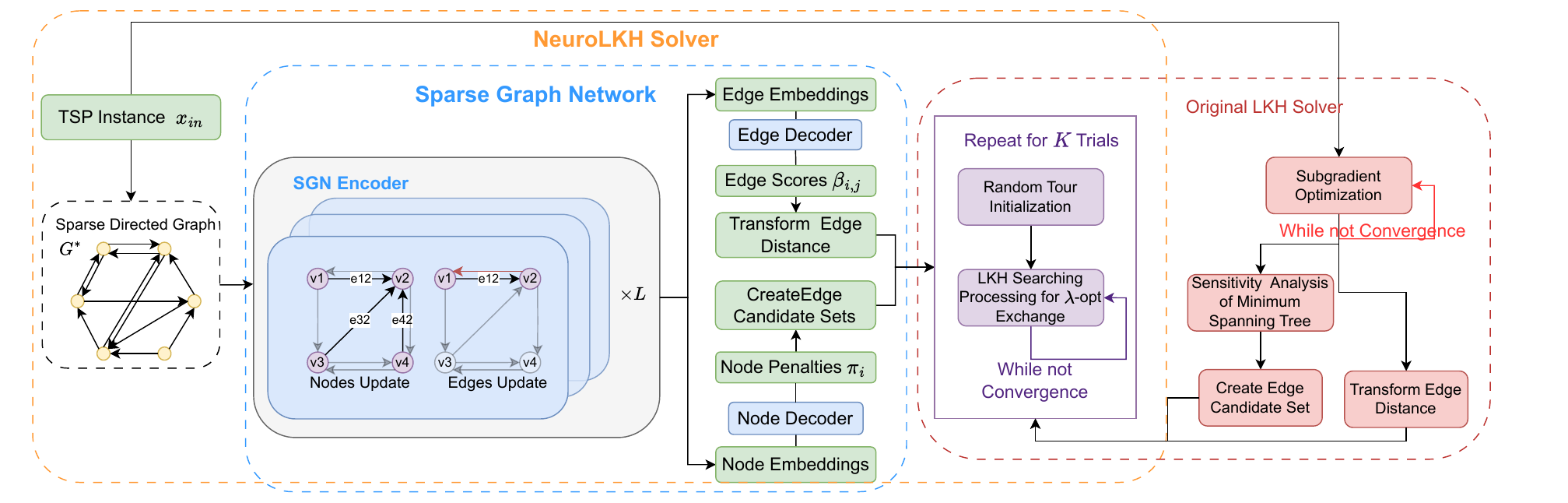} % Reduce the figure size so that it is slightly narrower than the column.
\caption{The overall structure of NLKH.}
\label{NLKH_Structure}
\end{figure*}

In the EAX algorithm, the selection strategies used during different stages of the algorithm are crucial for constructing the E-set, which is key to generating offspring. 
\begin{itemize}
    \item In Stage 1, the \textbf{Single Strategy} is applied. This strategy involves selecting a single $AB$-cycles randomly from the pool of $AB$-cycles generated by the parent solutions. The selection is done without overlapping previous selections, leading to the formation of an $E$-set that typically includes smaller cycles and more localized changes to the parent solution. This strategy is effective in generating offspring that are close to the parent solution, which helps in fine-tuning high-quality solutions early in the search process.
    \item In Stage 2, the \textbf{Block2 Strategy} is employed, which focuses on constructing an E-set with relatively few and large segments of the parent solutions ($p_A$ and $p_B$). This strategy aims to reduce the number of subtours in the intermediate solutions, thereby increasing the likelihood of generating high-quality offspring. The Block2 Strategy involves a tabu search process to efficiently explore combinations of $AB$-cycles, ensuring that the resulting $E$-set is effective in producing improved solutions. The detailed mechanism of the Block2 Strategy, including the use of central $AB$-cycles and the calculation of $\#C$ values to minimize subtours, is crucial in the later stages of the search, where more significant structural changes are necessary to escape local optima.
\end{itemize}

\section{The NLKH Method}

The NLKH solver integrates deep learning with the classical Lin-Kernighan-Helsgaun (LKH) heuristic to solve the Traveling Salesman Problem (TSP). By using a Sparse Graph Network (SGN), NLKH predicts node penalties and edge scores, refining the candidate set used in LKH and guiding the optimization process. As shown in Fig.~\ref{NLKH_Structure}, the overall structure of the NLKH solver is as follows:
\begin{itemize}
\item First, the TSP instance is represented as a graph, and a Sparse Graph Network (SGN) is employed to predict node penalties and edge scores.
\item The candidate set is constructed using the edge scores predicted by the SGN, where only the most promising edges are included.
\item The LKH algorithm is then applied, using the learned node penalties and candidate set to guide the search process.
\end{itemize}

\subsection{Sparse Graph Network (SGN)}
\subsubsection{SGN Encoder: }
The SGN Encoder takes as input a sparse directed graph  where $V$ is the set of nodes and $E^*$ is a sparse set of directed edges containing only the $\gamma$ shortest edges from each node. The node inputs $x_v \in \mathbb{R}^2$ (node coordinates) and edge inputs $x_e \in \mathbb{R}$ (edge distances) are first embedded into feature vectors through a linear projection, producing initial node features $v^0_i \in \mathbb{R}^D$ and edge features $e^0_{i,j} \in \mathbb{R}^D$, where $D$ is the feature dimension.
The node and edge features are then refined using $L$ Sparse Graph Convolutional Layers, which apply attention mechanisms and element-wise operations to update the embeddings. The node and edge embeddings are computed as:
\begin{equation}\label{Aeq1}
    a_{i, j}^l=\exp \left(W_a^l e_{i, j}^{l-1}\right) \oslash \sum\limits_{(i, m) \in E *} \exp \left(W_a^l e_{i, m}^{l-1}\right)
\end{equation}
\begin{equation}\label{Aeq6}
    \textstyle
    v_{i}^l = \mathcal{F}\left(W_s^l v_i^{l-1} +  \sum\limits_{(i,j)\in E^*} a_{i,j}^l \odot W_n^lv_j^{l-1}\right)
    +v_{i}^{l-1}
\end{equation}
\begin{equation}\label{Aeq7}
    r_{i, j}^l= \begin{cases}W_r^l e_{j, i}^{l-1}, & \text { if }\left(j, i\right) \in E^* \\ W_r^l p^l, & \textrm{ otherwise }\end{cases}
\end{equation}
\begin{equation}
    \textstyle
    e_{i,j}^l = \mathcal{F}\left(W_f^l v_i^{l-1} + W_t^lv_j^{l-1}+W_o^{l}e_{i,j}^{l-1}+r_{i,j}^l\right)+e_{i,j}^{l-1},
\end{equation}
where $\odot$ and $\oslash$ represent element-wise multiplication and element-wise division, respectively; $l = 1, 2, ..., L$ is the layer index;
$W_a^l$, $W_n^l$, $W_s^l$, $W_r^l$, $W_f^l$, $W_t^l$, $W_o^l \in \mathbb{R}^{D \times D}$ are trainable parameters; $\mathcal{F}$ represents ReLU activation followed by Batch Normalization.

\subsubsection{SGN Decoder: }
In NLKH, the node penalties $\pi_i$ are computed by passing the final node features $v^L_i$ through two layers of linear projections followed by a ReLU activation. The resulting values are then scaled using a tanh function, ensuring that the penalties are bounded within a specific range, mathematically represented as follows: 
\begin{equation}\label{Aeq8_unique}
    \textstyle
   \pi_i = C \cdot \tanh(W_\pi v^L_i),
\end{equation}
where $W_\pi \in \mathbb{R}^{D}$ is a trainable weight matrix and $C = 10$ is used to keep the node penalties in the range of [-10, 10].

Based on $e^L_{i,j}$ output by the encoder, the decoder generates the final embedding vectors $e^F_{i,j}$ using two layers of linear projection and ReLU activation.
Then, the edge score $\beta_{i,j}$ is calculated as follows: 
\begin{equation}\label{A:eq9}
    \beta_{i,j} = \dfrac{\exp\left(W_{\beta}e_{i,j}^{F}\right)}{\sum_{\left(i,m\right)\in E^{*}}\exp\left(W_{\beta} e_{i,m}^{F}\right)}.
\end{equation}
where $W_\beta \in \mathbb{R}^{D}$ is a trainable parameter and $E^*$ denotes the edges in the sparse graph. This process assigns a probability score to each edge, indicating its likelihood of being part of the optimal tour. 
\subsubsection{Training Procedure}
The training process of NLKH involves both supervised and unsupervised learning components to optimize the edge scores and node penalties, respectively. The edge scores $\beta_{i,j}$ are trained using a supervised learning approach, where a cross-entropy loss is minimized. This loss function is defined as follows:
\begin{equation}
    \begin{aligned}
        \mathcal{L}_\beta = -\frac{1}{\gamma \left\vert V\right\vert} \sum\limits_{(i,j)\in E^*} ( & \mathbb{I}\left\{(i,j) \in E_o^*\right\}\log\left(\beta_{i,j}\right) \\
                                                                                                   & + \mathbb{I} \left\{(i,j) \notin E_o^*\right\} \log (1-\beta_{i,j}))
    \end{aligned}
\end{equation}
where $E^*o$ represents the edges in the optimal tour. On the other hand, the node penalties $\pi_i$ are optimized using unsupervised learning by minimizing the degree deviation in the Minimum 1-Tree, with the loss function as follows:
\begin{equation}\label{eq10}
    \begin{aligned}
        L_\pi = -\frac{1}{|V|} \sum_{i \in V} (d_i(\pi) - 2) \pi_i,
    \end{aligned}
\end{equation}
where $d_i(\pi)$ is the degree of node $i$ in the Minimum 1-Tree. The overall loss function used in the training process is a weighted sum of these two components, expressed as $L = L_\beta + \eta_\pi L_\pi$, where $\eta_\pi$ is a balancing coefficient that adjusts the contribution of the node penalties to the total loss. This combined approach allows the model to learn both the edge scores and node penalties effectively, which are crucial for guiding the LKH algorithm.

\subsection{The Original LKH Method}
The LKH method optimizes TSP solutions through iterative \( \lambda \)-opt moves, exchanging \( \lambda \) edges to reduce the tour length. It uses a precomputed candidate set, traditionally derived from Minimum Spanning Tree analysis, to direct the search. NLKH improves this by using a Sparse Graph Network to learn and generate a more effective candidate set, reducing search time and improving solution quality.

\section{Comparison with Deep Learning-based Methods}

\subsection{TSP Benchmarks}
Due to the instance size-dependent GPU memory consumption and inference time of LEHD, 33 instances smaller than 10,000 nodes were selected from the TSPLib, National, and VLSI benchmark sets. These instances were divided into two subsets: the Small set (15 instances, 3,500–5,000 nodes) and the Medium set (18 instances, 5,000–10,000 nodes). Due to the non-configurable and relatively short runtime nature of LEHD, only the Local Search (LS) phase was tested for comparison, denoted as UNiCS-LS.

\subsection{Testing Method}
The same hardware configuration as in the main experiments was employed: two AMD EPYC 7713 CPUs and a NVIDIA A800 80GB. For each instance, all methods were executed with 10 independent runs, following the protocol of the main experiments. To ensure fair comparisons, runtime parity between LEHD and UNiCS-LS was enforced. Because LEHD does not support explicit runtime configuration, the following procedure was adopted: LEHD was first executed to record its actual runtime per instance, and UNiCS-LS was then constrained to the identical runtime for that instance.

\begin{table}[h]
\centering
\caption{Relative gap (\%) to the BKS on the Large set (10 runs per instance). ``-'' in Best / Avg columns indicates no solutions produced in 10 runs / solutions produced only in some runs. Bold indicates best performance. ``Total'' row summarizes average gaps across Large set.} 
\label{Atable5}
\resizebox{0.8\columnwidth}{!}{
\begin{tabular}{ccccc}
\toprule
\multirow{2}{*}{Benchmark} & \multicolumn{2}{c}{UNiCS-LS} & \multicolumn{2}{c}{LEHD}  \\ \cmidrule(lr){2-3} \cmidrule(lr){4-5}
 & Best & Avg & Best & Avg \\ \midrule
{Small,(3500-5000]} & 0.1706 & 0.2848 & 22.4353 & 22.4353 \\
{Medium,(5000-10000]} & 0.0246 & 0.0689 & 62.6516 & 62.6516 \\
\bottomrule
\end{tabular}
}
\end{table}

\subsection{Results Analysis}
Table \ref{Atable5} summarizes the results on the Small and Medium datasets. LEHD, being a deterministic method, yields identical Best and Avg results. As shown, LEHD exhibits performance gaps of orders of magnitude compared to UNiCS-LS on both subsets, with even larger discrepancies on Medium. This indicates that neural network-based methods like LEHD remain inadequate for solving large-scale TSP instances derived from real-world scenarios. Notably, the Medium instances are still smaller than the smallest instances in our main experiments (Large set), justifying our omission of LEHD from the main text.  For more detailed results, refer to Table \ref{Atable6} and Table \ref{Atable7}.

While other deep learning-based methods (as reported in their papers) show no orders-of-magnitude differences from LEHD on TSPLib, our method achieves significantly better results than LEHD on Medium. These findings suggest that neural methods and our approach differ fundamentally in focus and performance: neural methods excel in small-scale instances (e.g., <1,000 nodes) with shorter runtimes but struggle with scalability. Thus, direct comparisons between our method and neural approaches are inappropriate, as their strengths lie in distinct problem regimes.

\newpage

\begin{table*}[h]
\centering
\caption{Relative gap (\%) to the BKS on the Small set (10 runs per instance). } 
\label{Atable6}
\resizebox{0.85\textwidth}{!}{
\begin{tabular}{|c|c|c|c|c|c|c|}
\hline
\multirow{2}*{Instance} & \multirow{2}*{Optimum} & \multirow{2}*{Run Time(s)} & \multicolumn{2}{c|}{UNiCS-LS} &  \multicolumn{2}{c|}{LEHD} \\
\cline{4-7}
 ~  &  ~  &  ~ & Best(Gap\%) & Average(Gap\%) & Best(Gap\%) & Average(Gap\%) \\
\hline
fjs3649 & 9272.0 & 39.69 & \textbf{9272.0(0.0000)} & \textbf{9273.4(0.0151)} & 12862.13281(38.7200) & 12862.13281(38.7200) \\
fjr3672 & 9601.0 & 40.31 & \textbf{9601.0(0.0000)} & \textbf{9601.8(0.0083)} & 12959.99414(34.9900) & 12959.99414(34.9900) \\
dlb3694 & 10959.0 & 41.27 & \textbf{10959.0(0.0000)} & \textbf{10960.4(0.0128)} & 12596.28711(14.9400) & 12596.28711(14.9400) \\
ltb3729 & 11821.0 & 41.75 & \textbf{11822.0(0.0085)} & \textbf{11824.6(0.0305)} & 13809.67871(16.8200) & 13809.67871(16.8200) \\
fl3795 & 28772.0 & 43.35 & \textbf{29465.0(2.4086)} & \textbf{29872.4(3.8246)} & 34838.125(21.0800) & 34838.125(21.0800) \\
xqe3891 & 11995.0 & 46.88 & \textbf{12000.0(0.0417)} & \textbf{12005.8(0.0900)} & 13756.92188(14.6900) & 13756.92188(14.6900) \\
\hline
\end{tabular}
}
\end{table*}

\begin{table*}[h]
\centering
\caption{Relative gap (\%) to the BKS on the Small set (10 runs per instance). } 
\label{Atable7}
\resizebox{0.85\textwidth}{!}{
\begin{tabular}{|c|c|c|c|c|c|c|}
\hline
\multirow{2}*{Instance} & \multirow{2}*{Optimum} & \multirow{2}*{Run Time(s)} & \multicolumn{2}{c|}{UNiCS-LS} &  \multicolumn{2}{c|}{LEHD}  \\
\cline{4-7}
 ~  &  ~  &  ~  & Best(Gap\%) & Average(Gap\%) & Best(Gap\%) & Average(Gap\%) \\ 
\hline
fqm5087 & 13029.0 & 85.25 & \textbf{13035.0(0.0461)} & \textbf{13032.0(0.0230)} & 18167.37891(39.4400) & 18167.37891(39.4400) \\
fea5557 & 15445.0 & 108.42 & \textbf{15449.6(0.0298)} & \textbf{15445.0(0.0000)} & 18359.70313(18.8700) & 18359.70313(18.8700) \\
rl5915 & 565530.0 & 127.29 & \textbf{565651.8(0.0215)} & \textbf{565615.0(0.0150)} & 672688.875(18.9500) & 672688.875(18.9500) \\
rl5934 & 556045.0 & 128.22 & \textbf{556595.8(0.0991)} & \textbf{556152.0(0.0192)} & 648011.4375(16.5400) & 648011.4375(16.5400) \\
tz6117 & 394718.0 & 136.84 & \textbf{394919.2(0.0510)} & \textbf{394751.0(0.0084)} & 599770.875(51.9500) & 599770.875(51.9500) \\
xsc6880 & 21535.0 & 187.09 & \textbf{21544.8(0.0455)} & \textbf{21543.0(0.0371)} & 25774.41016(19.6900) & 25774.41016(19.6900) \\
\hline
\end{tabular}
}
\end{table*}

\begin{table*}[tbp] 
\begin{center}
\begin{small}
\caption{Results on each TSP instance with Runtime = 1800s.}
\label{Atable1}
\resizebox{\textwidth}{!}{
\begin{tabular}{|c|c|c|c|c|c|c|c|c|c|c|c|}
\hline
\multirow{2}*{Instance} & \multirow{2}*{Optimum} & \multicolumn{2}{c|}{LKH-A} & \multicolumn{2}{c|}{LKH-P} & \multicolumn{2}{c|}{RHGA} & \multicolumn{2}{c|}{EAX} & \multicolumn{2}{c|}{UNiCS} \\
\cline{3-12}
 ~ &  ~ & Best(Gap\%) & Average(Gap\%) & Best(Gap\%) & Average(Gap\%) & Best(Gap\%) & Average(Gap\%) & Best(Gap\%) & Average(Gap\%) & Best(Gap\%) & Average(Gap\%) \\
\hline
xmc10150 & 28387 & 28389.00(0.0070) & 28394.10(0.0250) & 28388.00(0.0035) & 28390.10(0.0109) & \textbf{28387.00}(\textbf{0.0000}) & \textbf{28387.00}(\textbf{0.0000}) & \textbf{28387.00}(\textbf{0.0000}) & 28387.73(0.0026) & \textbf{28387.00}(\textbf{0.0000}) & \textbf{28387.00}(\textbf{0.0000})\\
fi10639 & 520527 & 520536.00(0.0017) & 520558.10(0.0060) & \textbf{520527.00}(\textbf{0.0000}) & 520554.10(0.0052) & \textbf{520527.00}(\textbf{0.0000}) & 520531.75(0.0009) & \textbf{520527.00}(\textbf{0.0000}) & \textbf{520528.18}(\textbf{0.0002}) & \textbf{520527.00}(\textbf{0.0000}) & 520530.18(0.0006)\\
rl11849 & 923288 & 923312.00(0.0026) & 923463.50(0.0190) & \textbf{923288.00}(\textbf{0.0000}) & 923365.90(0.0084) & 923300.00(0.0013) & 923304.75(0.0018) & \textbf{923288.00}(\textbf{0.0000}) & 923297.45(0.0010) & 923292.00(0.0004) & \textbf{923296.00}(\textbf{0.0009})\\
usa13509 & 19982859 & \textbf{19982949.00}(\textbf{0.0005}) & 19983785.80(0.0046) & 19983103.00(0.0012) & 19983995.50(0.0057) & 19984036.00(0.0059) & 19988941.25(0.0304) & 19983236.00(0.0019) & 19983469.82(0.0031) & 19983081.00(0.0011) & \textbf{19983331.00}(\textbf{0.0024})\\
xvb13584 & 37083 & \textbf{37083.00}(\textbf{0.0000}) & 37091.40(0.0227) & 37085.00(0.0054) & 37091.50(0.0229) & 37085.00(0.0054) & 37086.25(0.0088) & \textbf{37083.00}(\textbf{0.0000}) & \textbf{37083.55}(\textbf{0.0015}) & \textbf{37083.00}(\textbf{0.0000}) & 37084.45(0.0039)\\
\hline
\end{tabular}
}
\end{small}
\end{center}
% \vskip -0.1in
\end{table*}

\begin{table*}[tbp]
\begin{center}
\begin{small}
\caption{Continued results on each TSP instance with Runtime = 1800s.}
\resizebox{\textwidth}{!}{
\begin{tabular}{|c|c|c|c|c|c|c|c|c|c|c|c|}
\hline
\multirow{2}*{Instance} & \multirow{2}*{Optimum} & \multicolumn{2}{c|}{NLKH} & \multicolumn{2}{c|}{VSRLKH-A} & \multicolumn{2}{c|}{VSRLKH-P} & \multicolumn{2}{c|}{UNiCS-P} & \multicolumn{2}{c|}{UNiCS}  \\
\cline{3-12}
 ~ &  ~ & Best(Gap\%) & Average(Gap\%) & Best(Gap\%) & Average(Gap\%) & Best(Gap\%) & Average(Gap\%) & Best(Gap\%) & Average(Gap\%) & Best(Gap\%) & Average(Gap\%)  \\
 \hline
xmc10150 & 28387 & 28397.00(0.0352) & 28413.20(0.0923) & \textbf{28387.00}(\textbf{0.0000}) & 28389.56(0.0090) & \textbf{28387.00}(\textbf{0.0000}) & 28388.70(0.0060) & \textbf{28387.00}(\textbf{0.0000}) & \textbf{28387.00}(\textbf{0.0000}) & \textbf{28387.00}(\textbf{0.0000}) & \textbf{28387.00}(\textbf{0.0000})\\
fi10639 & 520527 & 520538.00(0.0021) & 520603.90(0.0148) & 520539.00(0.0023) & 520565.00(0.0073) & \textbf{520527.00}(\textbf{0.0000}) & 520543.20(0.0031) & 520528.00(0.0002) & \textbf{520528.50}(\textbf{0.0003}) & \textbf{520527.00}(\textbf{0.0000}) & 520530.18(0.0006)\\
rl11849 & 923288 & 923551.00(0.0285) & 924807.50(0.1646) & \textbf{923288.00}(\textbf{0.0000}) & 923322.00(0.0037) & \textbf{923288.00}(\textbf{0.0000}) & 923315.80(0.0030) & \textbf{923288.00}(\textbf{0.0000}) & 923306.33(0.0020) & 923292.00(0.0004) & \textbf{923296.00}(\textbf{0.0009})\\
usa13509 & 19982859 & 19995132.00(0.0614) & 20001844.70(0.0950) & 19983104.00(0.0012) & 19984129.78(0.0064) & 19983333.00(0.0024) & 19983860.50(0.0050) & \textbf{19982952.00}(\textbf{0.0005}) & \textbf{19982995.83}(\textbf{0.0007}) & 19983081.00(0.0011) & 19983331.00(0.0024)\\
xvb13584 & 37083 & 37103.00(0.0539) & 37108.40(0.0685) & 37085.00(0.0054) & 37092.56(0.0258) & 37085.00(0.0054) & 37091.60(0.0232) & 37085.00(0.0054) & 37085.00(0.0054) & \textbf{37083.00}(\textbf{0.0000}) & \textbf{37084.45}(\textbf{0.0039})\\
\hline
\end{tabular}
}
\end{small}
\end{center}
% \vskip -0.1in
\end{table*}

\begin{table*}[tbp] 
\begin{center}
\begin{small}
\caption{Results on each TSP instance with Runtime = 3600s.}
\resizebox{\textwidth}{!}{
\begin{tabular}{|c|c|c|c|c|c|c|c|c|c|c|c|}
\hline
\multirow{2}*{Instance} & \multirow{2}*{Optimum} & \multicolumn{2}{c|}{LKH-A} & \multicolumn{2}{c|}{LKH-P} & \multicolumn{2}{c|}{RHGA} & \multicolumn{2}{c|}{EAX} & \multicolumn{2}{c|}{UNiCS} \\
\cline{3-12}
 ~ &  ~ & Best(Gap\%) & Average(Gap\%) & Best(Gap\%) & Average(Gap\%) & Best(Gap\%) & Average(Gap\%) & Best(Gap\%) & Average(Gap\%) & Best(Gap\%) & Average(Gap\%) \\
\hline
xmc10150 & 28387 & \textbf{28387.00}(\textbf{0.0000}) & 28389.80(0.0099) & \textbf{28387.00}(\textbf{0.0000}) & 28388.40(0.0049) & \textbf{28387.00}(\textbf{0.0000}) & \textbf{28387.00}(\textbf{0.0000}) & \textbf{28387.00}(\textbf{0.0000}) & \textbf{28387.00}(\textbf{0.0000}) & \textbf{28387.00}(\textbf{0.0000}) & \textbf{28387.00}(\textbf{0.0000})\\
fi10639 & 520527 & 520531.00(0.0008) & 520548.50(0.0041) & \textbf{520527.00}(\textbf{0.0000}) & 520542.30(0.0029) & \textbf{520527.00}(\textbf{0.0000}) & 520529.75(0.0005) & \textbf{520527.00}(\textbf{0.0000}) & \textbf{520527.55}(\textbf{0.0001}) & \textbf{520527.00}(\textbf{0.0000}) & 520528.18(0.0002)\\
rl11849 & 923288 & 923312.00(0.0026) & 923395.10(0.0116) & \textbf{923288.00}(\textbf{0.0000}) & 923349.60(0.0067) & 923300.00(0.0013) & 923304.25(0.0018) & \textbf{923288.00}(\textbf{0.0000}) & 923297.45(0.0010) & \textbf{923288.00}(\textbf{0.0000}) & \textbf{923292.36}(\textbf{0.0005})\\
usa13509 & 19982859 & \textbf{19982874.00}(\textbf{0.0001}) & 19983411.50(0.0028) & 19983103.00(0.0012) & 19983818.40(0.0048) & 19983023.00(0.0008) & 19983503.50(0.0032) & 19982910.00(0.0003) & \textbf{19983181.00}(\textbf{0.0016}) & 19983081.00(0.0011) & 19983255.73(0.0020)\\
xvb13584 & 37083 & \textbf{37083.00}(\textbf{0.0000}) & 37090.40(0.0200) & 37084.00(0.0027) & 37090.10(0.0191) & 37084.00(0.0027) & 37084.75(0.0047) & \textbf{37083.00}(\textbf{0.0000}) & \textbf{37083.45}(\textbf{0.0012}) & \textbf{37083.00}(\textbf{0.0000}) & 37083.91(0.0025)\\
\hline
\end{tabular}
}
\end{small}
\end{center}
\end{table*}

\begin{table*}[tbp] 
\begin{center}
\begin{small}
\caption{Continued results on each TSP instance with Runtime = 3600s.}
\label{Atable2}
\resizebox{\textwidth}{!}{
\begin{tabular}{|c|c|c|c|c|c|c|c|c|c|c|c|}
\hline
\multirow{2}*{Instance} & \multirow{2}*{Optimum} & \multicolumn{2}{c|}{NLKH} & \multicolumn{2}{c|}{VSRLKH-A} & \multicolumn{2}{c|}{VSRLKH-P} & \multicolumn{2}{c|}{UNiCS-P} & \multicolumn{2}{c|}{UNiCS}  \\
\cline{3-12}
 ~ &  ~ & Best(Gap\%) & Average(Gap\%) & Best(Gap\%) & Average(Gap\%) & Best(Gap\%) & Average(Gap\%) & Best(Gap\%) & Average(Gap\%) & Best(Gap\%) & Average(Gap\%)  \\
 \hline
xmc10150 & 28387 & 28396.00(0.0317) & 28405.60(0.0655) & \textbf{28387.00}(\textbf{0.0000}) & 28387.89(0.0031) & \textbf{28387.00}(\textbf{0.0000}) & 28387.80(0.0028) & \textbf{28387.00}(\textbf{0.0000}) & \textbf{28387.00}(\textbf{0.0000}) & \textbf{28387.00}(\textbf{0.0000}) & \textbf{28387.00}(\textbf{0.0000})\\
fi10639 & 520527 & 520538.00(0.0021) & 520585.00(0.0111) & 520535.00(0.0015) & 520552.22(0.0048) & \textbf{520527.00}(\textbf{0.0000}) & 520539.10(0.0023) & 520528.00(0.0002) & 520528.50(0.0003) & \textbf{520527.00}(\textbf{0.0000}) & \textbf{520528.18}(\textbf{0.0002})\\
rl11849 & 923288 & 923551.00(0.0285) & 924521.80(0.1336) & \textbf{923288.00}(\textbf{0.0000}) & 923297.44(0.0010) & \textbf{923288.00}(\textbf{0.0000}) & 923296.70(0.0009) & \textbf{923288.00}(\textbf{0.0000}) & \textbf{923289.58}(\textbf{0.0002}) & \textbf{923288.00}(\textbf{0.0000}) & 923292.36(0.0005)\\
usa13509 & 19982859 & 19991234.00(0.0419) & 19997697.50(0.0743) & 19983104.00(0.0012) & 19983527.00(0.0033) & 19983235.00(0.0019) & 19983488.10(0.0031) & \textbf{19982952.00}(\textbf{0.0005}) & \textbf{19982995.83}(\textbf{0.0007}) & 19983081.00(0.0011) & 19983255.73(0.0020)\\
xvb13584 & 37083 & 37096.00(0.0351) & 37102.20(0.0518) & 37085.00(0.0054) & 37090.78(0.0210) & 37084.00(0.0027) & 37089.30(0.0170) & 37084.00(0.0027) & 37084.00(0.0027) & \textbf{37083.00}(\textbf{0.0000}) & \textbf{37083.91}(\textbf{0.0025})\\
\hline
\end{tabular}
}
\end{small}
\end{center}
\end{table*}

\begin{table*}[tbp] 
\begin{center}
\begin{small}
\caption{Results on each instance with Runtime = 7200s.}
\label{Atable3}
% \vskip 0.1in
\resizebox{\textwidth}{!}{
\begin{tabular}{|c|c|c|c|c|c|c|c|c|c|c|c|}
\hline
\multirow{2}*{Instance} & \multirow{2}*{Optimum} & \multicolumn{2}{c|}{LKH-A} & \multicolumn{2}{c|}{LKH-P} & \multicolumn{2}{c|}{RHGA} & \multicolumn{2}{c|}{EAX} & \multicolumn{2}{c|}{UNiCS}  \\
\cline{3-12}
 ~ &  ~ & Best(Gap\%) & Average(Gap\%) & Best(Gap\%) & Average(Gap\%) & Best(Gap\%) & Average(Gap\%) & Best(Gap\%) & Average(Gap\%) & Best(Gap\%) & Average(Gap\%)  \\
 \hline
xmc10150 & 28387 & \textbf{28387.00}(\textbf{0.0000}) & 28389.20(0.0078) & \textbf{28387.00}(\textbf{0.0000}) & 28387.70(0.0025) & \textbf{28387.00}(\textbf{0.0000}) & \textbf{28387.00}(\textbf{0.0000}) & \textbf{28387.00}(\textbf{0.0000}) & \textbf{28387.00}(\textbf{0.0000}) & \textbf{28387.00}(\textbf{0.0000}) & \textbf{28387.00}(\textbf{0.0000})\\
fi10639 & 520527 & 520531.00(0.0008) & 520541.20(0.0027) & \textbf{520527.00}(\textbf{0.0000}) & 520531.00(0.0008) & \textbf{520527.00}(\textbf{0.0000}) & 520529.50(0.0005) & \textbf{520527.00}(\textbf{0.0000}) & \textbf{520527.18}(\textbf{0.0000}) & \textbf{520527.00}(\textbf{0.0000}) & 520527.82(0.0002)\\
rl11849 & 923288 & \textbf{923288.00}(\textbf{0.0000}) & 923355.50(0.0073) & \textbf{923288.00}(\textbf{0.0000}) & 923295.20(0.0008) & 923292.00(0.0004) & 923300.00(0.0013) & \textbf{923288.00}(\textbf{0.0000}) & 923292.36(0.0005) & \textbf{923288.00}(\textbf{0.0000}) & \textbf{923288.00}(\textbf{0.0000})\\
usa13509 & 19982859 & \textbf{19982874.00}(\textbf{0.0001}) & 19983095.20(0.0012) & 19982875.00(0.0001) & 19983240.60(0.0019) & 19982979.00(0.0006) & 19983087.50(0.0011) & 19982905.00(0.0002) & \textbf{19982928.00}(\textbf{0.0003}) & 19983045.00(0.0009) & 19983111.64(0.0013)\\
xvb13584 & 37083 & \textbf{37083.00}(\textbf{0.0000}) & 37088.30(0.0143) & 37084.00(0.0027) & 37088.30(0.0143) & 37084.00(0.0027) & 37084.75(0.0047) & \textbf{37083.00}(\textbf{0.0000}) & \textbf{37083.27}(\textbf{0.0007}) & \textbf{37083.00}(\textbf{0.0000}) & \textbf{37083.27}(\textbf{0.0007})\\
\hline
\end{tabular}
}
\end{small}
\end{center}
% \vskip -0.1in
\end{table*}

\begin{table*}[tbp] 
\begin{center}
\begin{small}
\caption{Continued results on each instance with Runtime = 7200s.}
\label{Atable4}
% \vskip 0.1in
\resizebox{\textwidth}{!}{
\begin{tabular}{|c|c|c|c|c|c|c|c|c|c|c|c|}
\hline
\multirow{2}*{Instance} & \multirow{2}*{Optimum} & \multicolumn{2}{c|}{NLKH} & \multicolumn{2}{c|}{VSRLKH-A} & \multicolumn{2}{c|}{VSRLKH-P} & \multicolumn{2}{c|}{UNiCS-P} & \multicolumn{2}{c|}{UNiCS}  \\
\cline{3-12}
 ~ &  ~ & Best(Gap\%) & Average(Gap\%) & Best(Gap\%) & Average(Gap\%) & Best(Gap\%) & Average(Gap\%) & Best(Gap\%) & Average(Gap\%) & Best(Gap\%) & Average(Gap\%)  \\
 \hline
xmc10150 & 28387 & 28396.00(0.0317) & 28403.20(0.0571) & \textbf{28387.00}(\textbf{0.0000}) & 28387.89(0.0031) & \textbf{28387.00}(\textbf{0.0000}) & 28387.40(0.0014) & \textbf{28387.00}(\textbf{0.0000}) & \textbf{28387.00}(\textbf{0.0000}) & \textbf{28387.00}(\textbf{0.0000}) & \textbf{28387.00}(\textbf{0.0000})\\
fi10639 & 520527 & 520538.00(0.0021) & 520568.40(0.0080) & \textbf{520527.00}(\textbf{0.0000}) & 520536.00(0.0017) & \textbf{520527.00}(\textbf{0.0000}) & 520533.10(0.0012) & \textbf{520527.00}(\textbf{0.0000}) & 520528.17(0.0002) & \textbf{520527.00}(\textbf{0.0000}) & \textbf{520527.82}(\textbf{0.0002})\\
rl11849 & 923288 & 923508.00(0.0238) & 923923.70(0.0689) & \textbf{923288.00}(\textbf{0.0000}) & \textbf{923288.00}(\textbf{0.0000}) & \textbf{923288.00}(\textbf{0.0000}) & 923290.40(0.0003) & \textbf{923288.00}(\textbf{0.0000}) & \textbf{923288.00}(\textbf{0.0000}) & \textbf{923288.00}(\textbf{0.0000}) & \textbf{923288.00}(\textbf{0.0000})\\
usa13509 & 19982859 & 19991086.00(0.0412) & 19995303.40(0.0623) & 19983029.00(0.0009) & 19983372.56(0.0026) & 19983062.00(0.0010) & 19983359.30(0.0025) & \textbf{19982952.00}(\textbf{0.0005}) & \textbf{19982989.25}(\textbf{0.0007}) & 19983045.00(0.0009) & 19983111.64(0.0013)\\
xvb13584 & 37083 & 37095.00(0.0324) & 37099.20(0.0437) & 37084.00(0.0027) & 37088.44(0.0147) & 37084.00(0.0027) & 37086.60(0.0097) & 37084.00(0.0027) & 37084.00(0.0027) & \textbf{37083.00}(\textbf{0.0000}) & \textbf{37083.27}(\textbf{0.0007})\\
\hline
\end{tabular}
}
\end{small}
\end{center}
% \vskip -0.1in
\end{table*}

\end{document}